\documentclass[10pt,twocolumn,letterpaper]{article}

\usepackage[pagenumbers]{cvpr} 

\usepackage[dvipsnames]{xcolor}

\newcommand{\myparagraph}[1]{\noindent\textbf{#1}\:}

\definecolor{cvprblue}{rgb}{0.21,0.49,0.74}
\usepackage[pagebackref,breaklinks,colorlinks,citecolor=cvprblue]{hyperref}
\usepackage{multirow}
\usepackage{makecell}
\usepackage{caption}
\usepackage{rotating}

\def\paperID{10498} 
\def\confName{CVPR}
\def\confYear{2024}

\title{PartSLIP++: Enhancing Low-Shot 3D Part Segmentation via Multi-View Instance Segmentation and Maximum Likelihood Estimation}

\author{
Yuchen Zhou\footnotemark[1]\thanks{\,Equal contributions; Corresponding Authors: yuz256@ucsd.edu, \\ jigu@ucsd.edu}\qquad
Jiayuan Gu\footnotemark[1]\qquad
Xuanlin Li\qquad
Minghua Liu\qquad
Yunhao Fang\qquad
Hao Su \qquad 
\vspace{0.1cm} \\
UC San Diego
}

\begin{document}
\maketitle
\begin{abstract}
Open-world 3D part segmentation is pivotal in diverse applications such as robotics and AR/VR. Traditional supervised methods often grapple with limited 3D data availability and struggle to generalize to unseen object categories. PartSLIP, a recent advancement, has made significant strides in zero- and few-shot 3D part segmentation. This is achieved by harnessing the capabilities of the 2D open-vocabulary detection module, GLIP, and introducing a heuristic method for converting and lifting multi-view 2D bounding box predictions into 3D segmentation masks. In this paper, we introduce PartSLIP++, an enhanced version designed to overcome the limitations of its predecessor. Our approach incorporates two major improvements. First, we utilize a pre-trained 2D segmentation model, SAM, to produce pixel-wise 2D segmentations, yielding more precise and accurate annotations than the 2D bounding boxes used in PartSLIP. Second, PartSLIP++ replaces the heuristic 3D conversion process with an innovative modified Expectation-Maximization algorithm. This algorithm conceptualizes 3D instance segmentation as unobserved latent variables, and then iteratively refines them through an alternating process of 2D-3D matching and optimization with gradient descent. Through extensive evaluations, we show that PartSLIP++ demonstrates better performance over PartSLIP in both low-shot 3D semantic and instance-based object part segmentation tasks. We finally showcase the versatility of PartSLIP++ in enabling applications like semi-automatic part annotation and 3D Instance Proposal Generation. Code released at \url{https://github.com/zyc00/PartSLIP2}.
\end{abstract}    
\section{Introduction}
\label{sec:intro}


3D part segmentation focuses on dividing a 3D shape into distinct parts, which necessitates a comprehensive understanding of the object's structure, semantics, mobility, and functionality. It plays a crucial role in various applications, including robotics, AR/VR, and shape analysis and synthesis~\cite{aleotti20123d,liu2022frame,xu2022unsupervised,mo2019structurenet}. 

Remarkable progress has been made in developing diverse data-driven approaches for 3D part segmentation~\cite{qian2022pointnext,wang2019dynamic,liu2019relation,yi2019gspn}. However, standard supervised training necessitates a substantial volume of finely-annotated 3D training shapes, the collection and annotation of which are typically labor-intensive and time-consuming. For instance, the PartNet dataset~\cite{mo2019partnet}, which is the most extensive publicly available 3D part dataset, comprises 26,000 objects but covers only 24 common everyday categories. Such limited training categories often hinders supervised methods from effectively tackling open-world scenarios and handling out-of-distribution test shapes (e.g., unseen classes).

Contrary to 3D data, 2D images accompanied by text descriptions are more readily available, contributing significantly to the recent advancements in large-scale image-language models~\cite{radford2021learning,jia2021scaling,li2022grounded,zhang2022glipv2,alayrac2022flamingo,ramesh2022hierarchical,saharia2022photorealistic}. A recent work, PartSLIP~\cite{liu2023partslip}, thus capitalizes on this by utilizing the rich 2D priors and the robust zero-shot capabilities of the image-language model to address the 3D part segmentation task in a zero or few-shot fashion. PartSLIP begins by rendering multi-view images for an input 3D point cloud. These images, along with a text prompt, are fed into the GLIP~\cite{li2022grounded} model, known for its proficiency in open-world 2D detection. To translate the 2D bounding boxes detected by GLIP into 3D semantic and instance segmentation masks, PartSLIP introduces a heuristic pipeline involving superpoint generation, 3D voting, and 3D grouping. While PartSLIP has shown impressive zero-shot and few-shot performance, it does have some notable drawbacks: \textbf{(a)} the 2D bounding boxes generated by GLIP can be coarse, lacking pixel-level accurate part annotations; \textbf{(b)} the heuristic pipeline might not yield the most accurate 3D segmentation; \textbf{(c)} the heuristic relies on multiple hyperparameters, making the final results sensitive to their specific settings.

In this work, we propose \textbf{PartSLIP++}, a novel method designed to surpass the aforementioned limitations and further enhance its performance. This method primarily incorporates two significant modifications. Firstly, we generate pixel-wise 2D annotations by utilizing a pre-trained 2D segmentation model, SAM~\cite{kirillov2023segany}. Specifically, SAM uses initially-detected bounding boxes from GLIP as prompts to generate precise 2D instance segmentations. These pixel-wise segmentation masks offer more accurate 2D annotations compared to the bounding boxes used in the prior work, PartSLIP. Secondly, rather than relying on a heuristic lifting algorithm in PartSLIP, we formulate the conversion from multi-view 2D segmentation to 3D segmentation as a problem of maximum likelihood estimation with latent variables. To address this, we introduce a modified EM algorithm~\cite{dempster1977maximum}. Here, the 3D instance segmentation mask is treated as an unobserved latent variable. During the E-step, the Hungarian algorithm is applied to match the predicted 2D instance segmentation masks with the current estimate of projected 3D instance segmentation masks, aiming to calculate the expectation of the log-likelihood. In the M-step, the 3D instance segmentation is updated by minimizing a cost function based on the matches established in the E-step. This algorithm iteratively alternates between these two steps until convergence is reached.

In our comprehensive evaluation using the PartNetE dataset~\cite{liu2023partslip}, we demonstrate that PartNet++ outperforms PartSLIP in terms of both low-shot 3D semantic and instance segmentation tasks. Additionally, our detailed ablation studies highlight the effectiveness of each module and design technique we propose. Key contributions of our work include:

\begin{itemize}
\setlength\itemsep{0.4em}
    \item Integrating a pre-trained 2D segmentation model into the PartSLIP pipeline, yielding more accurate and precise 2D pixel-wise part annotations than the bounding boxes used in prior work.
    \item Reformulating the problem of lifting multi-view 2D part segmentation masks to 3D masks as a maximum likelihood estimation problem, and introducing a novel modified Expectation-Maximization (EM) algorithm for effective optimization of this problem.
    \item Demonstrating that PartSLIP++ outperforms existing low-shot baselines in both 3D semantic and instance-based part segmentation through quantitative and qualitative analysis. The effectiveness of PartSLIP++ further enables applications like semi-automatic part annotation and 3D Instance Proposal Generation.
\end{itemize}

\section{Related Works}
\label{sec:related-works}

\subsection{3D Part Segmentation}

There are two main tasks for 3D segmentation: semantic and instance segmentation. Semantic segmentation is to predict a semantic label for each geometric primitive (e.g., point~\cite{qi2017pointnet}, voxel~\cite{graham20183d}, superpoints~\cite{landrieu2018large}).
For learning-based instance segmentation, there are mainly two lines of works: bottom-up and top-down approaches. 
The bottom-up approaches~\cite{jiang2020pointgroup,liu2020self,wang2018sgpn,zhang2021point,he2020learning,vu2022softgroup,wang2019associatively,chu2021icm} usually learn instance-aware features and cluster geometric primitives into different instances based on the distance metric defined on those features.
The top-down approaches~\cite{yi2019gspn,yang2019learning,hou20193d} usually first generate region proposals and then segment the foreground within each region of interest.
Recently, transformers~\cite{vaswani2017attention} are also introduced for 3D instance segmentation~\cite{schult2023mask3d,lu2023query}. Each object instance is represented as an instance query, and a transformer decoder is applied to predict instance masks.

Most works above address scene-level 3D semantic segmentation and object-level 3D instance segmentation. Part-level 3D segmentation~\cite{yu2019partnet,luo2020learning,wang2021learning,bokhovkin2021towards,notchenko2022scan2part} has its unique challenges. For example, part instances are closer to each other and smaller than object instances. Besides, some part instances can be encompassed by other objects (e.g., a handle in the door). \cite{yu2019partnet} proposes a method that predicts a fixed number of part instance masks given a point cloud. During training, it uses the Hungarian algorithm to match each predicted instance mask with a ground-truth instance mask for supervision.

\subsection{Multi-view 2D-3D Segmentation}
Many works have studied how to tackle 3D understanding problems by multi-view approaches, e.g., shape classification~\cite{qi2016volumetric} and semantic segmentation~\cite{dai20183dmv,jaritz2019multi,mascaro2021diffuser}.
Given recent progress in 2D foundation models, several works have explored how to transfer the knowledge of 2D foundation models to 3D in a multi-view fashion.
PointCLIP~\cite{zhang2022pointclip} enables low-shot shape classification by aggregating the view-wise features of rendered multi-view depth maps encoded by CLIP~\cite{radford2021learning}.
LeRF~\cite{kerr2023lerf} distills CLIP features into a language embedded radiance field through NeRF-style optimization, which can support pixel-aligned, zero-shot queries.
In addition, SA3D~\cite{cen2023segment} generalizes a powerful vision foundation model SAM~\cite{kirillov2023segany} to segment 3D objects also via NeRF-style optimization.
Recently, PartSLIP~\cite{liu2023partslip} proposes a pipeline to tackle 3D part segmentation with the help of open-vocabulary 2D object detection models like GLIP~\cite{li2022grounded}, detailed in the next section.

\section{Method}

\begin{figure*}[t]
    \centering
    \includegraphics[width=\textwidth]{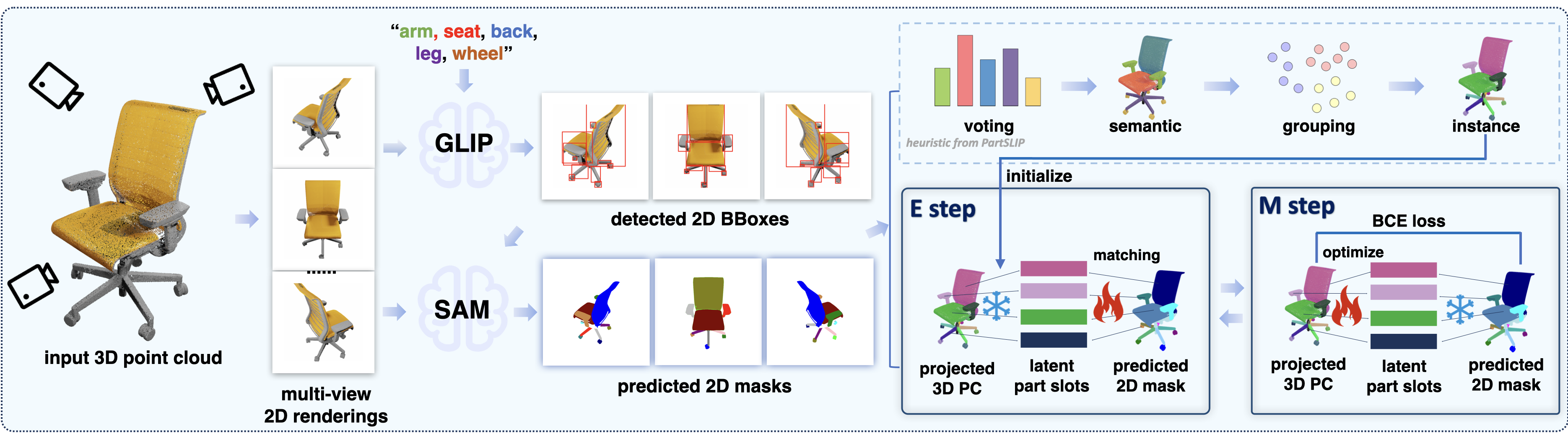}
    \caption{PartSLIP++ begins by taking a dense 3D point cloud as its input. It initially renders multi-view images from this point cloud. These images, along with a text prompt, are then input into the GLIP model, which predicts 2D bounding boxes. Subsequently, we utilize the SAM model to generate 2D instance segmentation masks for each view, using the predicted 2D bounding boxes as prompts. These multi-view 2D instance masks are converted into a 3D part segmentation mask using a novel, modified EM algorithm. During the E-step, the Hungarian algorithm is employed to find the optimal match between the projected 3D segmentation and the 2D predicted instance masks. In the M-step, the found matching is used to refine the 3D segmentation through gradient descent optimization. Lastly, the heuristic method presented by PartSLIP is applied to initialize the 3D instance segmentation.}
    \label{fig:overview}
\end{figure*}

We first review the prior work PartSLIP in Sec.~\ref{sec:partslip-review}. We refer readers to the original paper for more details. Then, we revisit the multi-view 2D-3D segmentation pipeline in Sec.~\ref{sec:revisit-mv-seg}, and propose a straightforward but effective way to improve 2D segmentation results, which can be a bottleneck for multi-view approaches.
Last, we propose a modified EM algorithm to merge multi-vew 2D segmentation results into 3D part labels in Sec.~\ref{sec:MLE}.
Fig.~\ref{fig:overview} provides an overview of our improved pipeline PartSLIP++.

\subsection{Preliminary: PartSLIP}
\label{sec:partslip-review}
\cite{liu2023partslip} introduces a pipeline called PartSLIP, which leverages GLIP \cite{li2022grounded}, a pretrained open-vocabulary object detection model, to tackle both semantic and instance segmentation tasks for 3D object parts.
Given a colored point cloud, PartSLIP first renders multiple images from $K$ predefined camera poses. Then, each rendered image and a text prompt concatenating all part names of interest and the object category are fed into the GLIP model, which will predict multiple 2D bounding boxes for all part instances visible from the current view. Finally, all 2D bounding boxes from different views are merged into 3D part segmentation labels.
\cite{liu2023partslip} proposes a learning-free module to lift the 2D GLIP predictions to 3D part segmentation labels, which mainly contains 3 following components: 

\myparagraph{3D Superpoint Generation}
The input point cloud $P$ is first oversegmented into a collection of superpoints \cite{landrieu2018large} $\{SP_i\}$. Points in each superpoint share similar normals and colors, and are assumed to belong to the same instance. Part labels will be calculated based on superpoints instead of points, which can save much computation and lead to potentially better performance due to the 3D prior.

\myparagraph{3D Semantic Voting}
The semantic label of each superpoint is voted by all 2D bounding boxes from multiple views that overlap with its 2D projection. Concretely, for a superpoint $SP_i$ and a part category $j$, a score $s_{i,j}$ is calculated based on the ratio of visible points covered by 2D detected instances of the part category in each view:
\begin{equation}
\label{eq:semseg_score}
s_{i,\,j} = \frac{\sum_k \sum_{p \in SP_i} [\operatorname{VIS}_k(p)] [\exists b\in \mathcal{B}_k^j : \operatorname{INS}_b(p)]}{\sum_k \sum_{p \in SP_i} [\operatorname{VIS}_k(p)]}
\end{equation}
where $[\cdot]$ is the Iverson bracket (which evaluates to 1 if the predicate inside it is true, and 0 if false); $\operatorname{VIS}_k(p)$ indicates whether the 3D point $p$ is visible in view $k$; $\mathcal{B}_k^j$ is a set of predicted bounding boxes of category $j$ in view $k$; and $\operatorname{INS}_b(p)$ indicates whether the projection of point $p$ in view $k$ is inside the bounding box $b$. The part category with the highest score is assigned to the superpoint as its semantic label. 

\myparagraph{3D Instance Grouping}
\label{sec:partslip-instance-grouping}
PartSLIP~\cite{liu2023partslip} heuristically groups oversegmented superpoints into instances according to their semantic similarity, spatial adjacency and 2D label consistency across views. Specifically, two superpoints $SP_u$ and $SP_v$ are considered to belong to the same instance if (a) they share the same semantic label, (b) they are neighbors in a KNN graph, and (c) the overlaps between their 2D projections and detected 2D bounding boxes are similar in each view.
The overlap between the 2D projection of a superpoint $SP_u$ and a 2D bounding box $b \in \mathcal{B}_k$ in view $k$ is:
\begin{equation}
\label{eq:overlap_sp_box}
    o(SP_u,b) = \frac{\sum_{p\in SP_u} [\operatorname{VIS}_{k}(p)][\operatorname{INS}_{b}(p)]}{\sum_{p\in SP_u}[\operatorname{VIS}_{k}(p)]}
\end{equation}
\cite{liu2023partslip} considers a list of 2D bounding boxes $\mathcal{B}'$ from views where both of superpoints $SP_u$ and $SP_v$ are visible, and constructs two feature vectors $I_u, I_v \in \mathbb{R}^{|\mathcal{B}'|}$, where $I_u[i]=o(SP_u, \mathcal{B}'[i])$. The last criterion is satisfied if $\frac{|I_u-I_v|_1}{max(|I_u|_1, |I_v|_1)}$ is smaller then a predefined threshold.
3D instances can be found via the Union-Find algorithm.

\subsection{Revisiting Multi-view 2D-3D Segmentation}
\label{sec:revisit-mv-seg}
In this section, we will revisit how 3D segmentation is tackled by multi-view 2D segmentation. 
Given a colored point cloud $P$, the goal of 3D segmentation is to predict its label $Y$.
For multi-view 2D-3D segmentation approaches, with $K$ views rendered from the point cloud, a 2D instance segmentation model is first employed to generate instance segmentation masks $\mathcal{M}_k$ for each view $k$. The key is to merge 2D segmentation results from multiple views. This problem can be formulated as estimating the parameters $Y$ by maximizing the likelihood of $P(\{\mathcal{M}_k\}|Y)$. In other words, we try to find a 3D label assignment that is compatible with observed 2D predictions. Intuitively, if two points belong to the same predicted 2D instance in each view, chances are that they belong to the same 3D instance.

Due to the lack of strong open-vocabulary instance segmentation models at that time, PartSLIP resorted to the open-vocabulary object detection model GLIP, and used bounding boxes as coarse instance masks. However, a bounding box can cover irrelevant pixels from other instances, resulting in noisy 2D instance labels.
To address this issue, \textbf{we propose to convert GLIP to an open-vocabulary instance segmentation model by using a promptable 2D instance segmentation model to further segment instances within detected bounding boxes}. In this work, we use the Segment Anything Model (SAM)\cite{kirillov2023segany}. The predicate $\operatorname{INS}_b(p)$ in Eq.~\ref{eq:semseg_score} and \ref{eq:overlap_sp_box}, which indicates whether a point is inside a bounding box, can be replaced with $\operatorname{INS}_M(p)$, where $M$ is the instance mask output by SAM with the bounding box $b$ as the prompt.

Besides, PartSLIP does not directly maximize the likelihood of predicted 3D part instances. As mentioned in Sec. \ref{sec:partslip-instance-grouping}, it uses the Union-Find algorithm to group superpoints into instances based on distances between heuristically-designed features.
Such method can be sensitive to the threshold of feature distance to consider whether two superpoints can be merged.
To this end, \textbf{given multi-view 2D instance segmentations and initial 3D instances produced by the PartSLIP pipeline, we further refine these 3D instances by proposing a modified expectation-maximization (EM) algorithm to find the maximum-likelihood estimates of 3D instances}, detailed in the next section.

\subsection{2D-3D Part Segmentation with EM Algorithm}
\label{sec:MLE}

\subsubsection{Problem Definition}
Formally, we define the problem of multi-view 2D-3D segmentation as estimating the label $Y \in \mathbb{L}^{n}$ of a colored point cloud $P \in \mathbb{R}^{n \times 3}$ by maximizing the likelihood of $P(\mathcal{M}|Y)$, where $\mathcal{M}=\cup_{k=1}^K \mathcal{M}_k$ is the union of all predicted 2D instance masks from all $K$ views and $\mathcal{M}_k$ is the set of 2D instance masks in view $k$.
Here, $n$ is the number of points and $\mathbb{L}$ is a predefined set of labels. $\mathbb{L}$ is usually defined as a set of integers, the number of which is either the number of semantic categories for semantic segmentation, or the maximum number of instances for instance segmentation. We denote the number of labels by $l=|\mathbb{L}|$.

Without loss of generality, we take instance segmentation for example in this section. We introduce a parameter (3D instance labels) matrix $\Theta \in \mathbb{R}^{n \times l}$, where the i-th row $\Theta_{i,:}$ is the logit of the i-th point for 3D instance label and $Y_i=argmax_j(\Theta_{i,j})$.
Besides, we introduce a latent (2D-3D assignment) matrix $Z \in \{0, 1\}^{m \times l}$, where $m=|\mathcal{M}|$ is the total number of 2D predicted instances across views. $Z_{i,j}=1$ and $Z_{i,\neq j}=0$ indicate that the i-th 2D predicted instance should belong to the j-th 3D instance $j$.
The maximum likelihood estimate (MLE) of the unknown parameters $\Theta$ is determined by maximizing the marginal likelihood of the observed data $\mathcal{M}$:
\begin{equation}
\begin{split}
    &L(\Theta; \mathcal{M}) = P(\mathcal{M}|\Theta) \\= &\int_{Z} P(\mathcal{M},Z|\Theta) = \int_Z P(\mathcal{M}|Z,\Theta)P(Z|\Theta)
\end{split}
\label{eq:mle-em}
\end{equation}

To find the MLE of 3D instance label parameter $\Theta$, we apply the classical expectation-maximization (EM) algorithm~\cite{dempster1977maximum} with some modifications.
The EM algorithm is an iterative method, consisting of two steps at each EM iteration. An EM iteration alternates between performing an expectation (E) step to build a log likelihood function of parameters using the current estimate, and a maximization (M) step to find the parameters that maximize the likelihood function built in the E step.
In this work, we randomly select a view to perform updates at each EM iteration. In the E step (Sec.~\ref{sec:e-step}), we define a cost function (equivalent to a log likelihood function) to match each 2D predicted instance in the selected view with one of 3D instance labels, and update the latent 2D-3D assignment matrix $Z^{t+1}$ with the minimum total cost. In the M step (Sec.~\ref{sec:m-step}), we update the parameter matrix to $\Theta^{t+1}$ via minimizing the total cost in the E step by gradient descent.
The above problem definition and algorithm also apply to labeling superpoints.

\subsubsection{E Step: Matching 2D and 3D Instances}
\label{sec:e-step}

In the E step, we aim to match 2D predicted instances with 3D instance labels and induce a log likelihood function of 3D instance logits $\Theta$. Given the current estimate $\Theta^t$ and the instance segmentation masks $\mathcal{M}_k$ in the selected view k, we can define a cost function for a 2D-3D assignment $Z$. 
First, for each 3D instance label $j$, we denote a function $\Pi_k$ to project its scores $\hat{\Theta}_{:,j}$ to a 2D image $\Pi_k(\hat{\Theta}_{:,j}) \in \mathbb{R}^{H \times W}$, where the score $\hat{\Theta}_{i,:}$ of the i-th point is induced by applying a softmax function to the logit ${\Theta}_{i,:}$, and $H, W$ are the image height and width.
Next, we denote the i-th 2D instance mask in view $k$ by $\mathcal{M}_k^i \in \{0,1\}^{H \times W}$.
Then, if the i-th 2D instance is assigned with a 3D instance label $j$, the cost function is defined as negative log-likelihood:
\begin{equation}
\begin{split}
    C(\Pi_k(\hat{\Theta}_{:,j}), \mathcal{M}_k^i) = - \sum_{q} \Bigl( \mathcal{M}_k^i[q] log \Pi_k(\hat{\Theta}_{:,j})[q] \\ + (1-\mathcal{M}_k^i[q]) log (1 - \Pi_k(\hat{\Theta}_{:,j})[q]) \Bigr)
\end{split}
\label{eq:cost}
\end{equation}
where $q$ is a pixel position on the image. Given the cost function defined in Eq.~\ref{eq:cost}, we use the Hungarian Algorithm to find the optimal assignment $Z^{t+1}$.

\subsubsection{M Step: Optimizing 3D Instance Logits}
\label{sec:m-step}

In the M step, we can update the 3D instance logits $\Theta$ by minimizing the overall cost function $L(\Theta)$ given the assignment $Z^{t+1}$ found in the E step. We use the gradient descent to update the parameters.
\begin{equation}
    L(\Theta) = \sum_{i,j} [Z_{i,j}=1] C(\Pi_k(\hat{\Theta}_{:,j}), \mathcal{M}_k^i)
\end{equation}

\subsubsection{Initialization}

The EM algorithm can only find a local minimal, and a good initialization can typically lead to better solutions. Therefore, we use the 3D instance segmentation results from a pretrained PartSLIP checkpoint (introduced in Sec.~\ref{sec:partslip-instance-grouping}) to initialize $\Theta^0$.
Assume that $\hat{m} \leq l$ instances are found by grouping superpoints in PartSLIP. For the i-th point and the 3D instance label $j \in \{1,\dots,\hat{m}\}$, we have $\Theta^{0}_{i,j}=\log \hat{m}$ while $\Theta^{0}_{i,\neq j}=0$. 

\subsubsection{Post-processing}
\label{sec:post-processing}
A single 3D part instance is typically spatially adjacent, i.e., all points in a single instance form a single cluster based on spatial proximity. Our initial analysis finds that a 3D instance mask produced by our EM algorithm could sometimes contain multiple, disconnected instances. Therefore, we propose to further postprocess our 3D instances by splitting among them. Specifically, for each 3D instance, we use a spatial cluster algorithm similar to \cite{jiang2020pointgroup} to obtain one or more disjoint clusters among this instance. When an instance divides into multiple clusters, each becomes a separate 3D instance, retaining the original semantic label.

\section{Experiments}
\label{sec:eperiments}
\begin{table*}[t]
  \centering
  \scriptsize
  \setlength{\tabcolsep}{1.55pt}
  \caption{Semantic segmentation mIoU results on the PartNetE Dataset. We present results for the 17 object categories that overlap between PartNetE and PartNet, where in addition to the 8 training shapes from PartNetE per category, some baseline models also include an extra 28,000 shapes from PartNet, resulting in a total of 45x8+28k configurations. We also present results for the 28 unique categories in PartNetE, where models are trained using 8 PartNetE shapes from each category. For a detailed breakdown of performance on all 45 categories, please refer to the supplementary material. }
    \begin{tabular}{p{0.075\linewidth}|c|cccccccc|c|cccccccc|c|c}
    \toprule
    \multirow{3}[4]{*}{\#3D data} & \multirow{3}[4]{*}{method} & \multicolumn{9}{c|}{Overlapping Categories}                           & \multicolumn{9}{c|}{Non-Overlapping Categories}                       &  \\
\cmidrule{3-20}          &       & \multirow{2}[2]{*}{Bottle} & \multirow{2}[2]{*}{Chair} & \multirow{2}[2]{*}{Display} & \multirow{2}[2]{*}{Door} & \multirow{2}[2]{*}{Knife} & \multirow{2}[2]{*}{Lamp} & Storage & \multirow{2}[2]{*}{Table} & Overall & \multirow{2}[2]{*}{Camera} & \multirow{2}[2]{*}{Cart} & Dis-  & \multirow{2}[2]{*}{Kettle} & Kitchen- & \multirow{2}[2]{*}{Oven} & Suit- & \multirow{2}[2]{*}{Toaster} & Overall & Overll \\
          &       &       &       &       &       &       &       & Furniture &       & (17)  &       &       & Penser &       & Pot   &       & case  &       & (28)  & (45) \\
    \midrule
    \multirow{3}[2]{*}{\makecell{Few-shot w/ \\ extra data \\ (45x8+28k)}} & PointNet++~\cite{qi2017pointnet++} & 48.8  & 84.7  & 78.4  & 45.7  & 35.4  & 68.0  & 46.9  & \textbf{63.7}  & 55.6  & 6.5   & 6.4   & 12.1  & 20.9  & 15.8  & 34.3  & 40.6  & 14.7  & 25.4  & 36.8 \\
          & PointNext~\cite{qian2022pointnext} & 68.4  & \textbf{91.8}  & \textbf{89.4}  & 43.8  & 58.7  & 64.9  & \textbf{68.5}  & 52.1  & \textbf{58.5}  & 33.2  & 36.3  & 26.0  & 45.1  & 57.0  & 37.8  & 13.5  & 8.3   & \textbf{45.1}  & \textbf{50.2} \\
          & SoftGroup~\cite{vu2022softgroup} & 41.4  & 88.3  & 62.1  & \textbf{53.1}  & 31.3  & \textbf{82.2}  & 60.2  & 54.8  & 50.2  & 23.6  & 23.9  & 18.9  & 57.4  & 45.5  & 13.6  & 18.3  & 26.4  & 30.7  & 38.1  \\
    \midrule
    \multirow{6}[1]{*}{\makecell{Few-shot \\ (45x8)}} & PointNet++~\cite{qi2017pointnet++} & 27.0  & 42.2  & 30.2  & 20.5  & 22.2  & 10.5  & 8.4   & 7.3   & 18.1  & 9.7   & 11.6  & 7.0   & 28.6  & 31.7  & 19.4  & 3.3   & 0.0   & 21.8  & 20.4 \\
          & PointNext~\cite{qian2022pointnext} & 67.6  & 65.1  & 53.7  & 46.3  & 59.7  & 55.4  & 20.6  & 22.1  & 39.2  & 26.0  & 47.7  & 22.6  & 60.5  & 66.0  & 36.8  & 14.5  & 0.0   & 41.5  & 40.6 \\
          & SoftGroup~\cite{vu2022softgroup} & 20.8  & 80.5  & 39.7  & 16.3  & 38.3  & 38.3  & 18.9  & 24.9  & 32.8  & 28.6  & 40.8  & 42.9  & 60.7  & 54.8  & 35.6  & 29.8  & 14.8  & 41.1  & 38.0 \\
        & ACD~\cite{gadelha2020label}   & 22.4  & 39.0  & 29.2  & 18.9  & 39.6  & 13.7  & 7.6   & 13.5  & 19.2  & 10.1  & 31.5  & 19.4  & 40.2  & 51.8  & 8.9   & 13.2  & 0.0   & 25.6  & 23.2 \\
          & Prototype~\cite{zhao2021few} & 60.1  & 70.8  & 67.3  & 33.4  & 50.4  & 38.2  & 30.2  & 25.7  & 41.1  & 32.0  & 36.8  & 53.4  & 62.7  & 63.3  & 36.5  & 35.5  & 10.1  & 46.3  & 44.3 \\
          & PartSLIP~\cite{liu2023partslip}  & 83.4  & 85.3  & 84.8  & 40.8  & 65.2  & 66.0  & 53.6  & 42.4  & 56.3  & 58.3  & 88.1  & 73.7  & 77.0  & 69.6  & 73.5  & 70.4  & 60.0  & 61.3  & 59.4 \\
          & PartSLIP*~\cite{liu2023partslip}  & 81.2  & 82.7  & 81.8  & 43.1  & 62.5  & 66.3  & 52.3  & 44.3  & 56.6  & 61.8  & 79.0  & 71.0  & 73.3  & 66.5  & 69.1  & 64.5  & 50.1  & 58.7  & 57.9 \\
          & \textbf{Ours} & \textbf{85.8} & 85.3 & 85.1 & 45.1 & \textbf{64.3} & 67.9 & 57.2 & 45.3 & \textbf{57.0} & \textbf{63.2} & \textbf{84.8} & \textbf{72} & \textbf{85.6} & \textbf{76.8} & \textbf{70.3} & \textbf{70.0} & \textbf{50.7} & \textbf{63.3} & \textbf{60.8} \\
    \bottomrule
    \end{tabular}%
    \label{table:semseg}
\end{table*}%

\begin{table*}[t]
  \centering
  \scriptsize
  \setlength{\tabcolsep}{1.7pt}
  \caption{ Instance segmentation mAP@50 results on the PartNetE Dataset. For more comprehensive performance on all 45 categories, please refer to the supplementary material.}
    \begin{tabular}{p{0.06\linewidth}|c|cccccccc|c|cccccccc|c|c}
    \toprule
    \multirow{3}[4]{*}{\#3D data} & \multirow{3}[4]{*}{method} & \multicolumn{9}{c|}{Overlapping Categories}                           & \multicolumn{9}{c|}{Non-Overlapping Categories}                       &  \\
\cmidrule{3-20}          &       & \multirow{2}[2]{*}{Bottle} & \multirow{2}[2]{*}{Chair} & \multirow{2}[2]{*}{Display} & \multirow{2}[2]{*}{Door} & \multirow{2}[2]{*}{Knife} & \multirow{2}[2]{*}{Lamp} & Storage & \multirow{2}[2]{*}{Table} & Overall & \multirow{2}[2]{*}{Camera} & \multirow{2}[2]{*}{Cart} & Dis-  & \multirow{2}[2]{*}{Kettle} & Kitchen- & \multirow{2}[2]{*}{Oven} & Suit- & \multirow{2}[2]{*}{Toaster} & Overall & Overll \\
          &       &       &       &       &       &       &       & Furniture &       & (17)  &       &       & Penser &       & Pot   &       & case  &       & (28)  & (45) \\
    \midrule
    \multirow{2}[1]{*}{45x8+28k} & PointGroup~\cite{jiang2020pointgroup} &  38.2  & 87.6  & 65.1  & \textbf{23.4}  & 19.3  & 62.7  & \textbf{49.1}  & \textbf{46.4}  & 41.7  & 8.6   & 29.2  & 24.0  & 61.3  & 59.4  & 13.8  & 15.6  & 7.0   & 24.6  & 31.0 \\
          & SoftGroup~\cite{vu2022softgroup} & 43.9  & \textbf{89.1}  & 68.7  & 21.2  & 27.2  & 63.3  & \textbf{49.1}  & 46.2  & \textbf{42.4}  & 0.7   & 28.4  & 26.4  & 63.8  & 59.3  & 16.4  & 13.5  & 7.5   & \textbf{25.6}  & \textbf{31.9}  \\ \midrule
    \multirow{3}[0]{*}{\makecell{few-shot \\ (45x8)}} & PointGroup~\cite{jiang2020pointgroup} & 8.0   & 77.2  & 16.7  & 3.7   & 15.6  & 9.8   & 0.0   & 0.0   & 14.6  & 4.7   & 28.5  & 30.7  & 52.1  & 57.0  & 0.0   & 0.0   & 0.0   & 16.8  & 16.0 \\
          & SoftGroup~\cite{vu2022softgroup} & 22.4  & 87.7  & 27.5  & 5.6   & 10.3  & 19.4  & 11.6  & 14.2  & 21.3  & 11.2  & 29.8  & 37.8  & 63.4  & 65.7  & 10.4  & 8.0   & 10.7  & 28.4  & 25.7   \\
          & PartSLIP~\cite{liu2023partslip}  & 79.4  & 84.3  & 82.9  & 17.9  & 43.9  & 68.3  & 32.8  & 32.3  & 42.5  & 36.8  & \textbf{83.3}  & \textbf{63.5}  & 75.4  & 70.5  & \textbf{64.5}  & 44.9  & 38.4  & 46.2  & 44.8 \\ 
          & PartSLIP*~\cite{liu2023partslip}  & 74.4  & 79.3  & 64.2  & 14  & 43.3  & \textbf{69.5}  & 29.2  & 32.1  & 41.1  & 29.6  & 71  & 59.7  & 72.5  & 70.3  & 46.3  & 44.6  & 34.9  & 39.8  & 40.3 \\ 
          & \textbf{Ours}  & \textbf{78.5}  & 86.0  & \textbf{74.1}  & 17.6  & \textbf{46.0}  & 66.9  & 36.7  & 33.5  & \textbf{47.6}  & \textbf{29.7}  & 80.8  & 63.2  & \textbf{81.6}  & \textbf{80.7}  & 56.3  & \textbf{49.6}  & \textbf{41.5}  & \textbf{48.2}  & \textbf{48.0} \\ 
    \bottomrule
    \end{tabular}%
\label{table:inseg}
\end{table*}%

In this section, we provide quantitative and qualitative analysis to demonstrate the ability for PartSLIP++ to outperform existing few-shot baselines in both 3D semantic and instance-based part segmentation. Subsequently, we perform an ablation study to justify each design component of PartSLIP++. Beyond these evaluations, we also demonstrate the versatility of PartSLIP++ in two practical applications: semi-automatic annotation of 3D parts and 3D instance proposals generation.

\begin{figure*}[t]
    \vspace{-0.3em}
    \centering
    \includegraphics[width=\linewidth]{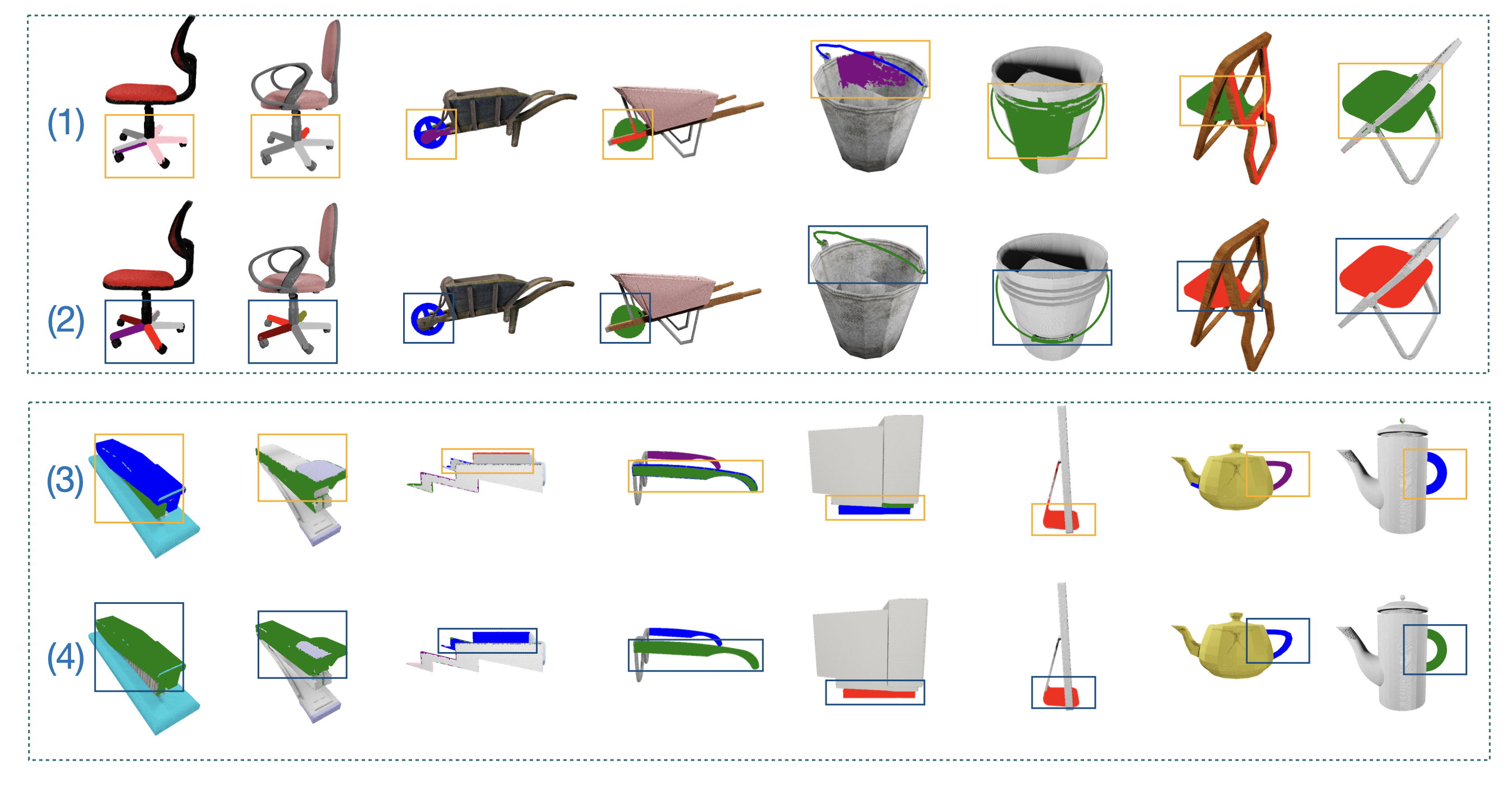}
    \caption{Qualitative analysis of 3D instance segmentation results for PartSLIP and PartSLIP++. Rows (1) and (3) illustrate the results from PartSLIP, and Rows (2) and (4) display the results from PartSLIP++. To enhance clarity, segmented instances are masked with a distinct color to differentiate from the object's original color, and are boxed to delineate the segmented areas. We find that in challenging tasks like segmenting thin bucket handles, the base of a computer monitor, or the seat of a swing chair, PartSLIP++ masks maintain a higher level of precision and adherence to the correct object parts, while PartSLIP masks often extend to undesired object areas.}
    \label{fig:real_world_demo}
    \vspace{1em}
\end{figure*}

\subsection{Datasets and Metrics}
\label{sec:dataset}
Following PartSLIP~\citep{liu2023partslip}, we adopt the PartNet-Ensemble (PartNet-E) dataset introduced in the paper, which consists of 1906 shapes covering 45 object categories, to evaluate our approach and the baselines. Our experiments encompass two settings: \textbf{(a)} Few-shot ($45 \times 8$): using 8 shapes for each of the 45 object categories. This setting is utilized in both our approach and the baseline. \textbf{(b)} Few shot with additional data ($45 \times 8 + 28k$): utilizing 28,367 shapes from PartNet\cite{mo2019partnet} (which has 17 categories that overlap with PartNet-E) in addition to the 45 × 8 shapes. This setting is only utilized in the baseline. We evaluate the semantic segmentation performance with mIoU and the instance segmentation performance with mAP@50.

\subsection{Implementation Details}
\label{sec:implmentation-details}

To ensure a fair comparison, we use the dataset released by PartSLIP~\cite{liu2023partslip}, which contains colored point clouds and camera poses used to render images. We follow the same setting to render each point cloud into 10 RGB images. We use the GLIP model finetuned on the low-shot data ($45 \times 8$), which is also released by PartSLIP. \textit{Note that the released checkpoint is known to have inferior performance compared to the version reported in the paper, confirmed by the authors of PartSLIP. We denote the released version by PartSLIP*.}

In our approach, to generate 2D instance masks given 2D detection results from GLIP, we utilize the pre-trained SAM~\cite{kirillov2023segany} model (ViT-H) without further task-specific fine-tuning, and use the detected bounding boxes as input prompts.
For the modified EM algorithm (Sec.~\ref{sec:MLE}), we use 10 EM iterations and the learning rate for gradient descent is 1.0 in the M step.
We adopt a threshold of 0.05 for the spatial clutering algorithm used in post-processing (Sec.~\ref{sec:post-processing}).

\subsection{Evaluation Results}
\label{sec:main-results}

We compare our PartSLIP++ with PartSLIP~\cite{liu2023partslip} on both semantic segmentation and instance segmentation tasks. For semantic segmentation, we additionally compare PartSLIP++ with PointNet++~\cite{qi2017pointnet++}, PointNext~\cite{qian2022pointnext}, and SoftGroup~\cite{vu2022softgroup}. For instance segmentation, we additionally compare PartSLIP++ with SoftGroup~\cite{vu2022softgroup} and PointGroup~\cite{jiang2020pointgroup}.

\myparagraph{Semantic Segmentation.} We present the semantic segmentation results in Table~\ref{table:semseg}. When training on the low-shot dataset of $45 \times 8$ shapes from PartNet-E, our PartSLIP++ attains the best performance compared to previous baselines. In particular, it outperforms released PartSLIP checkpoint by 2.9 mIoU (60.8 vs. 57.9) on the 45 categories in PartNet-E. The findings demonstrate PartSLIP++'s effectiveness in low-shot 3D semantic segmentation.

\myparagraph{Instance Segmentation.} We present the instance segmentation results in Table~\ref{table:inseg}. We find that our PartSLIP++ also achieves the best performance, with a notable 7.7 mAP improvement (48.0 vs. 40.3) over the released PartSLIP checkpoint. Furthermore, when evaluating PartSLIP++ on the 17 overlapping categories between PartNet-E and PartNet, even though PartSLIP++ is only trained on 8 shapes from each category, it outperforms the best model (SoftGroup) trained on an additional 28,000 shapes from the PartNet dataset by 5.2 mAP (47.6 vs. 42.4). The results demonstrate that PartSLIP++ is a strong model for low-shot 3D instance segmentation.

\myparagraph{Qualitative Analysis.} We present qualitative studies in Figure~\ref{fig:real_world_demo} to compare the 3D instance segmentation quality between PartSLIP++ and PartSLIP. Our observations reveal that PartSLIP++ excels in generating 3D instance masks that are more precise, accurate, and exhibit less noise. Notably, in challenging tasks like segmenting thin bucket handles, the base of a computer monitor, or the seat of a swing chair, PartSLIP++ demonstrates superior accuracy. The masks produced by PartSLIP often extend into undesired areas of the object, whereas those from PartSLIP++ maintain a higher level of precision and adherence to the correct object parts.

\subsection{Ablation Studies}
\label{sec:ablations}


\begin{table}[]
\centering
\small
\caption{Ablation study on the EM algorithm used in PartSLIP++. We report the mAP@50 for 3D instance segmentation on all the part categories. The results on three categories (chair, kettle, suitcase) are shown as well.}
\label{tab:ablation-all}
\begin{tabular}{c|ccc|c}
\hline
Method                & Chair & Kettle & Suitcase & Overall \\ 
\hline
PartSLIP++ (full)       & 86.0  & 81.6   & 49.6     & 48.0  \\
w/o post-processing    & 82.7  & 78.6   & 49.2     & 46.9  \\
w/o PartSLIP init   & 67.0    & 76.4   & 55.0     & 46.3  \\
w/o EM              & 80.4  & 79.6   & 44.1     & 44.8  \\
\hline
PartSLIP              & 79.3  & 72.5   & 44.6     & 40.3  \\ 
\hline
\end{tabular}
\end{table}

\myparagraph{Design Choices in EM algorithm.}
Table~\ref{tab:ablation-all} shows the ablation study on \textbf{3} design choices of our EM algorithm used in PartSLIP++: 1) whether to use the EM algorithm to refine initial 3D instance segmentations, 2) whether to initialize EM with 3D instance segmentations from PartSLIP, 3) whether to apply post-processing. We report the mAP@50 of different methods for 3D instance segmentation on different part categories.

We find that PartSLIP++ (full) outperforms PartSLIP++ (w/o EM) by 3.2 mAP (48.0 vs. 44.8), which demonstrates the effectiveness of our proposed modified EM algorithm in refining initial 3D instance masks. Besides, PartSLIP++ (full) outperforms PartSLIP++ (w/o PartSLIP init) by 1.7 mAP (48.0 vs. 46.3). This observation highlights the importance of the quality of 3D instance segmentation initialization in our EM algorithm. Additionally, PartSLIP++ (full) outperforms PartSLIP++ (w/o post-processing) by 1.1 mAP (48.0 vs. 46.9), illustrating that our 3D instance post-processing provides a helpful boost to the 3D instance segmentation performance. Therefore, all three components proposed in PartSLIP++ play a significant role to the overall improvement over PartSLIP.

\myparagraph{Refining 2D instance segmentations with SAM.} We then perform an ablation to investigate the effectiveness of our design to refine 2D instance segmentations with SAM. Results are shown in Table~\ref{tab:ablation-all}. We find that PartSLIP++ (w/o EM) outperforms PartSLIP by 4.5 mAP (44.8 vs. 40.3), demonstrating the large improvements brought by the more accurate 2D instance segmentation results with the help of the SAM model.

\myparagraph{Number of 2D Views.}
In our main experiments, we used 10 views to render each point cloud. In this ablation study, we investigate whether PartSLIP++ can benefit from more views that more comprehensively cover an object. We report the mAP@50 results for 3D instance segmentation on 3 part categories (Display, Door, Knife) in Table \ref{tab:ablation-num-views}. The results confirm that PartSLIP++ produces improved 3D instance segmentation masks when provided with a broader range of views of an object. Furthermore, the benefits become much more modest when the EM module, a key component of PartSLIP++, is removed. This indicates the crucial role of our EM module in maximizing the gains from additional input views.

\begin{table}[t]
\centering
\small
\caption{Ablation study on the number of 2D input views (our previous experiments used 10 input views). We report the mAP@50 metric for 3D instance segmentation on three part categories: display, door, knife.}
\label{tab:ablation-num-views}
\begin{tabular}{c|cccc}
\hline
Method & \begin{tabular}[c]{@{}c@{}}Number\\ of views\end{tabular} & Display & Door & Knife \\ \hline
       & 10                                                        & 74.1    & 17.6 & 46.0  \\
PartSLIP++  & 24                                                        & 77.8    & 24.8 & 51.1  \\
       & Gain                                                      & +3.7     & +7.2  & +5.1   \\ \hline
       & 10                                                        & 69.5    & 17.7 & 42.6  \\
PartSLIP++ w/o EM & 24                                                        & 71.4    & 19.9 & 44.2  \\
       & Gain                                                      & +1.9     & +2.2  & +1.6   \\ \hline
\end{tabular}
\end{table}

\subsection{Application: Part Annotation}
In this section, we illustrate the versatility of PartSLIP++ by illustrating its application in semi-automatic 3D object part annotation pipeline. In particular, PartSLIP++ is capable of segmenting 3D parts using multi-view 2D segmentation masks \textbf{without requiring the matching relationship between different views}. Based on this capability, we propose an annotation pipeline wherein annotators focus solely on labeling multi-view 2D images, assisted by the SegmentAnything. Once the multi-view images of a single object are fully annotated, our PartSLIP++ is automatically initiated in the backend. This process is designed to maximize efficiency in part annotation.

To test the robustness of this pipeline, we conduct a preliminary experiment. We randomly select several shapes from the PartNet-E dataset and manually annotate their 2D multi-view images. Subsequently, we independently apply the 3D instance mask generation pipelines in PartSLIP++ and PartSLIP to obtain 3D instance segmentations. Similar to our instance segmentation experiments, we employ mAP@50 as the evaluation metric. Results are shown in Table~\ref{tab:part_annotation_annotate}. To facilitate a comparison with human-annotated labels, we conduct an additional experiment where we condition PartSLIP++ and PartSLIP on multi-view ground-truth 2D segmentation masks. Results are presented in Table~\ref{tab:part_annotation_gt}. We find that for both human-annotated 2D masks and ground truth 2D masks, PartSLIP++ produces better 3D instance segmentations than PartSLIP, demonstrating the potential for PartSLIP to enhance the efficiency and accuracy of semi-automatic 3D object part annotation.

\begin{table}[]
\centering
\caption{mAP@50 results of 3D instance segmentation for PartSLIP++ and PartSLIP conditioned on multi-view manually-annotated 2D segmentations.}
\label{tab:my-table}
\begin{tabular}{c|cccc}
\hline
       method  & Chair & Suitcase & Knife \\ \hline
PartSLIP++ & 93.7      & 96.5         & 93.1     \\
PartSLIP & 88.1   &  97.3        & 84.5      \\ \hline
\end{tabular}%
\label{tab:part_annotation_annotate}
\end{table}
\begin{table}[]
\centering
\caption{mAP@50 results of 3D instance segmentation for PartSLIP++ and PartSLIP conditioned on multi-view ground-truth 2D segmentations.}
\label{tab:my-table}
\begin{tabular}{c|cccc}
\hline
        method & Chair  & Suitcase & Knife \\ \hline
PartSLIP++ & 99.6       & 100         & 94.0      \\
PartSLIP & 96.3       & 100         & 94.0      \\ \hline
\end{tabular}%
\label{tab:part_annotation_gt}
\end{table}

\subsection{Application: 3D Instance Proposal Generation}
In this section, we showcase class-agnostic 3D instance proposal generation powered by SAM and our modified EM algorithm. For many applications like part annotation, semantic information is not mandatory (or can be annotated easily), while the recall over part instances is critical. This motivates us to extend PartSLIP++ for class-agnostic 3D instance proposal generation.

\begin{figure}[t]
    \centering
    \includegraphics[width=\linewidth]{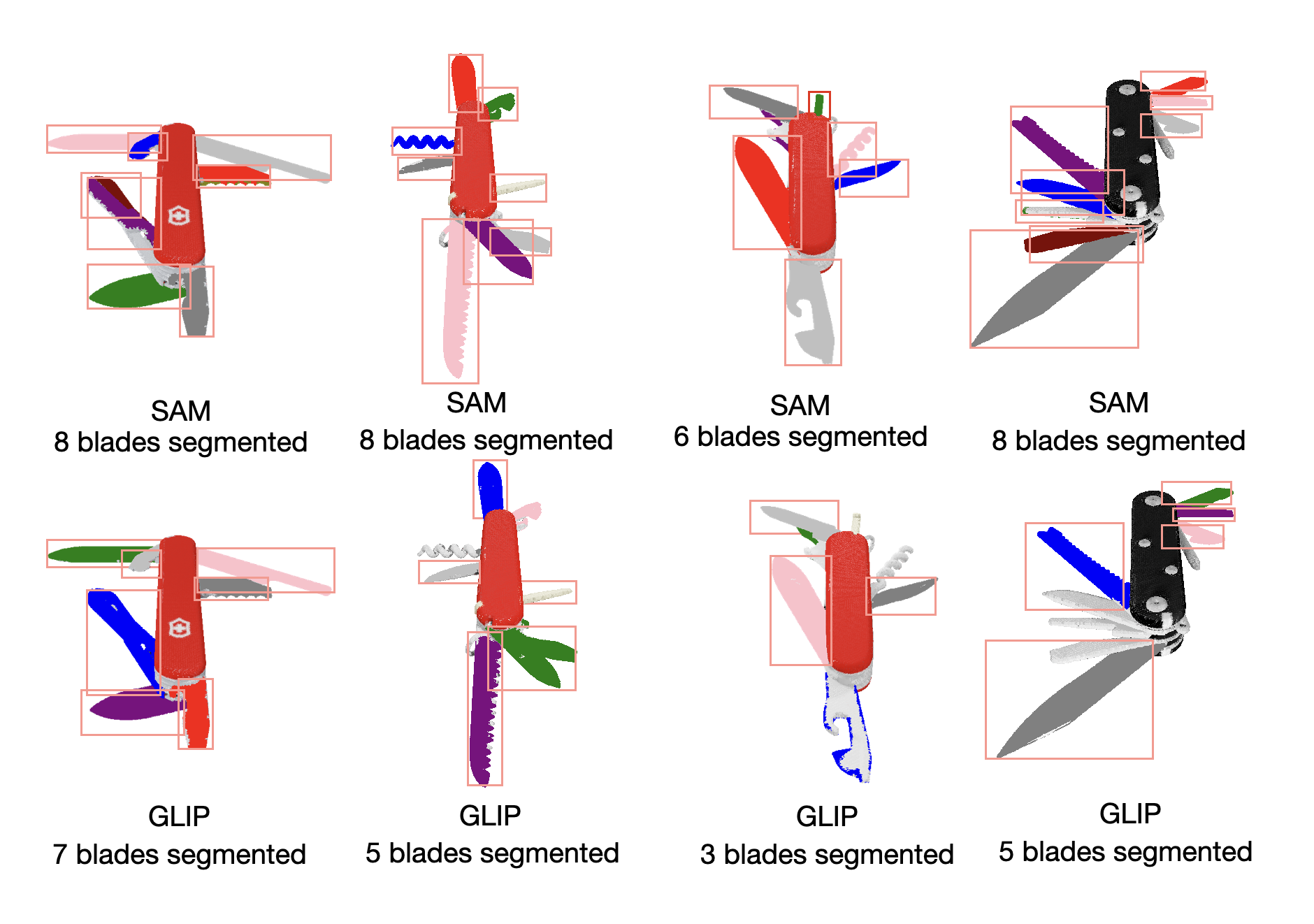}
    \caption{Example of 3D instance proposal generation. We extend PartSLIP++ by using SAM to directly generate class-agnostic instance proposals for each view and merging them with the modified EM algorithm. The first row shows the instance proposals generated by the (SAM-based) extension, and the second row shows the instances found by (GLIP-based) PartSLIP++. The number of blades segmented are shown below the visualization. The SAM-based extension shows a higher recall of part instances.}
    \label{fig:sam_everything}
\end{figure}

Concretely, we replace GLIP with SAM and leverage the ``segment everything'' ability of SAM to generate 2D instance proposals for each view. Then, our modified EM algorithm can be applied to merge 2D instance proposals from multiple views to 3D instance proposals.
Fig.~\ref{fig:sam_everything} showcases how this extension performs on \emph{knifes}, which contain many fine-grained parts (e.g., blades) that are especially challenging for open-vocabulary object detection models like GLIP. Compared to the GLIP-based PartSLIP++, the SAM-based extension yields more refined segmentation, as shown by the higher count of successfully segmented parts.

\section{Conclusion}

In this work, we propose PartSLIP++, a novel method for low-shot 3D semantic and instance segmentation on object parts that surpasses the limitations in the recent work PartSLIP. Specifically, PartSLIP++ first integrates a pre-trained 2D segmentation model to provide more accurate and precise 2D pixel-wise part annotations than the bounding boxes used in prior work. PartSLIP++ then formulates the problem of obtaining 3D instance segmentation from 2D multi-view instance labels as a maximum likelihood estimation problem, introducing a modified Expectation-Maximization (EM) algorithm for effective optimization. Through quantitative and qualitative analysis, we demonstrate that PartSLIP++ attains the best performance compared to previous approaches, and exhibits strong ability in low-shot 3D semantic and instance-based object part segmentation. We finally illustrate the versatility of PartSLIP++ in enabling diverse applications, such as semi-automatic part annotation and 3D instance proposal generation. 
{
    \small
    \bibliographystyle{ieeenat_fullname}
    \bibliography{main}
}

\clearpage


\setcounter{section}{0}
\renewcommand{\thesection}{\Alph{section}}
\section*{Appendix}
\section{Visualization of Part Annotation}

\begin{figure*}[t]
    \vspace{-0.3em}
    \centering
    \includegraphics[width=\linewidth]{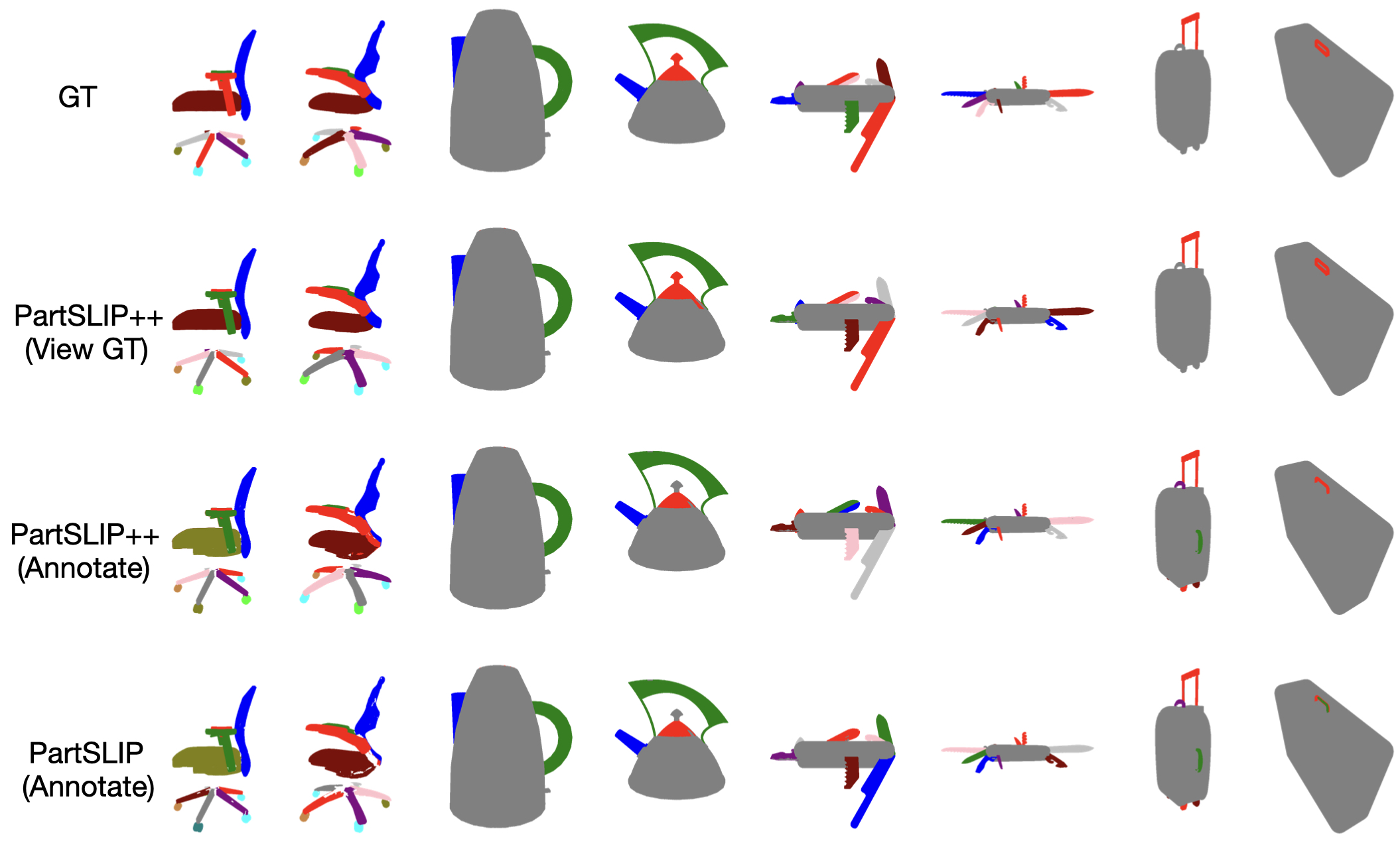}
    \caption{Qualitative analysis of the 3D part annotation application. The first row shows the ground truth 3D part segmentation labels. The second row shows our PartSLIP++'s 3D part segmentation result using  multi-view ground truth 2D segmentations as input. The third row shows our PartSLIP++'s 3D part segmentation result using human-annotated multi-view 2D segmentation masks as input. The forth row shows the baseline PartSLIP's 3D object part segmentation result using human-annotated multi-view 2D segmentation masks as input. By merging human-annotated multi-view results, PartSLIP++ can achieve 3D segmentation results close to groundtruth, which indicates the potential to annotate 3D part labels by multi-view annotations.}
    \label{fig:annotation-demo}
    \vspace{1em}
\end{figure*}

In this section, we provide a qualitative analysis of our application that uses PartSLIP++ to achieve semi-automatic 3D object part annotation. Specifically, after humans annotate multi-view 2D part segmentations, PartSLIP++ takes them as input to generate 3D part segmentations. The entire process can be achieved without knowing the matching relationship
between different views. Visualizations of 3D part segmentations generated on different objects are shown in Figure~\ref{fig:annotation-demo}. We find that compared to PartSLIP, PartSLIP++ is capable of generating masks with better precision and adherence to the correct object parts. The resulting 3D part segmentations are also closer to the ground-truth.

\section{Full Results on Semantic \& Instance Segmentation}

Tables \ref{tab:full_sem_0} and \ref{tab:full_sem_1} present the complete semantic segmentation mIoU results on all 45 categories of the PartNetE dataset. Table \ref{tab:ins_full} presents the complete instance segmentation results on all 45 categories of the PartNetE dataset.

\begin{table*}[t]
    \centering
    \tiny
    \setlength{\tabcolsep}{3.5pt}
    \setlength\extrarowheight{1pt}
    \caption{Full table (1/2) of semantic segmentation mIoU results on the PartNetE dataset. This table shows the results on 17 object categories that overlap between PartNetE and PartNet.}
      \begin{tabular}{c|cc|ccc|cccccccc}
      \toprule
      \multirow{53}[0]{*}{\begin{sideways}Overlapping Categories (17)\end{sideways}} & \multicolumn{2}{c|}{} & \multicolumn{3}{c|}{Few-shot w/ additional data (45x8+28k)} & \multicolumn{8}{c}{Few-shot (45x8)} \\
  \cmidrule{2-14}          & Category & Part  & \multicolumn{1}{c}{PointNet++~\cite{qi2017pointnet++}} & \multicolumn{1}{c}{PointNext~\cite{qian2022pointnext}} & \multicolumn{1}{c|}{SoftGroup~\cite{vu2022softgroup}} & \multicolumn{1}{c}{PointNet++~\cite{qi2017pointnet++}} & \multicolumn{1}{c}{PointNext~\cite{qian2022pointnext}} & \multicolumn{1}{c}{SoftGroup~\cite{vu2022softgroup}} & \multicolumn{1}{c}{ACD~\cite{gadelha2020label}} & \multicolumn{1}{c}{Prototype~\cite{zhao2021few}} & \multicolumn{1}{c}{PartSLIP~\cite{liu2023partslip}} & \multicolumn{1}{c}{PartSLIP*~\cite{liu2023partslip}} & \multicolumn{1}{c}{Ours} \\
  \cmidrule{2-14}          & Bottle & lid   & 48.8  & 68.4  & 41.4  & 27.0  & 67.6  & 20.8  & 22.4  & 60.1  & 83.4 & 80.8 & \textbf{85.5} \\
  \cline{2-14}          & \multirow{5}[0]{*}{Chair} & arm   & 83.5  & 88.6  & \textbf{89.7} & 29.5  & 68.6  & 67.8  & 27.6  & 58.7  & 74.1 & 65.4  & 69.6 \\
            &       & back  & 89.0  & \textbf{93.4} & 92.2  & 59.7  & 89.5  & 86.5  & 60.6  & 83.7  & 89.7 & 88.8  & 88.6 \\
            &       & leg   & 85.5  & \textbf{94.0} & 83.5  & 51.7  & 70.0  & 84.9  & 42.8  & 73.0  & 89.0 & 90.8  & 93.2 \\
            &       & seat  & 85.7  & \textbf{90.5} & 81.8  & 61.0  & 80.8  & 76.6  & 53.4  & 70.9  & 81.4 &  78.7  & 82.8 \\
            &       & wheel & 79.7  & 92.6  & \textbf{94.4} & 9.0   & 16.7  & 86.6  & 10.7  & 67.9  & 92.6 & 90.4  & 92.3 \\
  \cline{2-14}          & Clock & hand  & 19.2  & 28.4  & 2.5   & 0.0   & 0.0   & 6.0   & 0.0   & 10.5  & 37.6 & 39.2 & \textbf{54.1} \\
  \cline{2-14}          & \multirow{2}[0]{*}{Dishwasher} & door  & 59.3  & \textbf{81.5} & 50.7  & 55.6  & 73.9  & 54.2  & 50.6  & 68.6  & 71.2 & 68.8  & 71.1 \\
            &       & handle & 39.6  & \textbf{56.8} & 55.3  & 0.0   & 0.0   & 30.1  & 0.0   & 28.0  & 53.8 & 48.0  & 50.7 \\
  \cline{2-14}          & \multirow{3}[0]{*}{Display} & base  & 88.1  & 97.1 & 94.5  & 48.9  & 82.3  & 50.5  & 36.9  & 76.9  & 97.0 & 96.7  & \textbf{97.3} \\
            &       & screen & 80.4  & \textbf{87.6} & 49.6  & 40.1  & 78.8  & 46.1  & 42.1  & 73.6  & 73.9 & 68.7  & 75.5 \\
            &       & support & 66.5  & \textbf{83.4} & 42.3  & 1.5   & 0.0   & 22.6  & 8.4   & 51.5  & 83.4 & 77.4  & 82.6 \\
  \cline{2-14}          & \multirow{3}[0]{*}{Door} & frame & 48.2  & 50.0  & 42.6  & 22.6  & \textbf{65.6} & 23.4  & 23.5  & 49.1  & 20.9 & 19.5 & 17.7 \\
            &       & door  & 60.2  & \textbf{75.7} & 65.7  & 38.9  & 73.3  & 16.6  & 33.1  & 50.1  & 70.8 & 68.7 & 69.4 \\
            &       & handle & 28.6  & 5.7   & \textbf{51.0} & 0.0   & 0.0   & 8.9   & 0.0   & 1.2   & 30.7 & 41.7  & 48.5 \\
  \cline{2-14}          & \multirow{2}[0]{*}{Faucet} & spout & 80.1  & \textbf{90.4} & 82.6  & 31.2  & 67.2  & 50.4  & 31.4  & 62.1  & 79.0 & 75.2 & 75.7 \\
            &       & switch & 54.3  & \textbf{79.5} & 54.1  & 10.8  & 33.3  & 18.5  & 16.9  & 29.9  & 63.8 & 57.0 & 56.1 \\
  \cline{2-14}          & \multirow{2}[0]{*}{Keyboard} & cord  & 82.3  & 6.1   & 78.0  & 0.0   & 0.0   & 57.1  & 0.0   & 31.2  & 83.9 & 89.5 & \textbf{99.0} \\
            &       & key   & 66.7  & \textbf{83.8} & 39.8  & 31.5  & 69.2  & 50.2  & 52.2  & 58.5  & 23.3 & 56.2  & 45.7 \\
  \cline{2-14}          & Knife & blade & 35.4  & 58.7  & 31.3  & 22.2  & 59.7  & 38.3  & 39.6  & 50.4  & \textbf{65.2} & 62.3 & 64.3 \\
  \cline{2-14}          & \multirow{4}[0]{*}{Lamp} & base  & 77.5  & 72.8  & \textbf{92.8} & 20.5  & 82.0  & 48.7  & 6.0   & 56.2  & 90.3 & 88.3 & 89.2 \\
            &       & body  & 64.5  & 65.8  & 78.2  & 17.5  & 64.4  & 40.5  & 27.3  & 59.0  & 79.2 & 78.1 & \textbf{79.5} \\
            &       & bulb  & 51.4  & 35.2  & \textbf{66.3} & 0.0   & 0.0   & 12.2  & 0.0   & 4.4   & 10.2  & 12.5 & 13.5 \\
            &       & shade & 78.5  & 85.7  & \textbf{91.5} & 4.1   & 75.1  & 52.0  & 21.5  & 33.1  & 84.5  & 86.9 & 89.5 \\
  \cline{2-14}          & \multirow{5}[0]{*}{Laptop} & keyboard & 66.4  & \textbf{70.4} & 25.1  & 22.0  & 40.6  & 41.9  & 20.0  & 48.3  & 60.1 & 67.1 & 64.9 \\
            &       & screen & 79.0  & \textbf{83.0} & 33.9  & 28.4  & 79.9  & 42.6  & 35.5  & 68.2  & 62.8 & 57.7 & 60.5 \\
            &       & shaft & \textbf{27.7} & 0.0   & 19.6  & 0.0   & 0.0   & 13.4  & 0.0   & 8.7   & 3.0  & 3.0 & 2.1 \\
            &       & touchpad & \textbf{27.3} & 9.1   & 9.4   & 0.0   & 0.0   & 7.8   & 0.0   & 13.6  & 20.6  & 17.6 & 13.2 \\
            &       & camera & \textbf{76.6} & 0.0   & 4.1   & 0.0   & 0.0   & 0.9   & 0.0   & 0.7   & 2.1  & 14.5  & 7.5 \\
  \cline{2-14}          & \multirow{4}[0]{*}{Microwave} & display & 25.0 & 0.0   & 12.9  & 0.0   & 0.0   & 0.4   & 0.0   & 3.3   & 14.5 & \textbf{34.2} & 28.4 \\
            &       & door  & 63.6  & \textbf{75.4} & 44.9  & 25.0  & 63.9  & 51.8  & 26.5  & 62.0  & 45.2 & 40.3 & 52.5 \\
            &       & handle & 73.1  & 86.6  & 84.8  & 0.0   & 0.0   & 33.2  & 0.0   & 37.7  & \textbf{95.2} & 76.9 & 90.3 \\
            &       & button & 12.5  & 0.0   & 10.4  & 0.0   & 0.0   & 5.3   & 0.0   & 4.8   & 15.9 & 19.8 & \textbf{26.6} \\
  \cline{2-14}          & \multirow{2}[0]{*}{Refrigerator} & door  & 56.5  & \textbf{87.8} & 43.3  & 39.2  & 83.6  & 39.7  & 21.5  & 72.1  & 58.4 & 57.1 & 57.2 \\
            &       & handle & 30.3  & \textbf{64.5} & 50.4  & 0.0   & 0.0   & 31.0  & 0.0   & 13.6  & 53.1 & 47.7 & 54.1 \\
  \cline{2-14}          & \multirow{3}[0]{*}{Scissors} & blade & 59.0  & 82.1  & \textbf{85.2} & 44.5  & 72.7  & 74.0  & 52.6  & 45.4  & 76.8 & 73.1 & 72.6 \\
            &       & handle & 78.1  & 89.8  & \textbf{90.8} & 65.2  & 83.4  & 79.0  & 64.7  & 79.7  & 86.8 & 86.9 & 86.9 \\
            &       & screw & 12.8  & 0.0   & \textbf{52.0} & 0.0   & 0.0   & 14.0  & 0.0   & 3.9   & 17.4 & 22.7 & 21.9 \\
  \cline{2-14}          & \multirow{3}[0]{*}{StorageFurniture} & door  & 64.2  & \textbf{71.9} & 69.1  & 25.2  & 61.9  & 21.6  & 22.5  & 54.7  & 56.4 & 50.5 & 57.1 \\
            &       & drawer & 65.6  & \textbf{80.8} & 43.9  & 0.0   & 0.0   & 17.0  & 0.3   & 26.7  & 33.0 & 35.4 & 37.0 \\
            &       & handle & 10.9  & 52.8  & 67.6  & 0.0   & 0.0   & 18.0  & 0.0   & 9.2   & 71.4 & 70.8 & \textbf{77.8} \\
  \cline{2-14}          & \multirow{6}[0]{*}{Table} & door  & \textbf{71.7} & 14.5  & 33.6  & 0.0   & 0.0   & 0.0   & 0.0   & 0.0   & 0.0  & 0.0 & 0.0 \\
            &       & drawer & 42.3  & \textbf{55.6} & 41.0  & 8.3   & 35.0  & 29.1  & 22.0  & 24.9  & 35.3 & 36.8 & 36.6 \\
            &       & leg   & 67.3  & \textbf{85.0} & 64.4  & 15.8  & 15.4  & 45.7  & 17.7  & 53.7  & 66.4 & 70.9 & 72.0 \\
            &       & tabletop & 80.2  & \textbf{93.8} & 74.7  & 19.7  & 82.2  & 55.0  & 41.1  & 74.5  & 79.7 & 71.9 & 79.5 \\
            &       & wheel & 80.0  & 51.8  & 58.9  & 0.0   & 0.0   & 0.0   & 0.0   & 0.0   & 61.0 & \textbf{64.0} & 63.8 \\
            &       & handle & 40.9  & 11.8  & \textbf{56.3} & 0.0   & 0.0   & 19.4  & 0.0   & 1.2   & 12.3 & 20.1 & 20.3 \\
  \cline{2-14}          & \multirow{3}[0]{*}{TrashCan} & footpedal & \textbf{82.3} & 0.0   & 1.4   & 0.0   & 0.0   & 0.9   & 0.0   & 37.7  & 0.0  & 0.0 & 0.0 \\
            &       & lid   & 55.5  & \textbf{68.5} & 49.7  & 4.0   & 59.6  & 26.9  & 0.0   & 60.9  & 64.8 & 62.1 & 65.9 \\
            &       & door  & \textbf{77.4} & 0.0   & 0.0   & 0.9   & 0.0   & 0.0   & 0.0   & 0.0   & 2.1  & 8.4 & 6.7 \\
  \cmidrule{2-14}          & \multicolumn{2}{c|}{Overall (17)} & 55.6  & \textbf{58.5} & 50.2  & 18.1  & 39.2  & 32.8  & 19.2  & 41.1  & 56.3 & 56.6 & \textbf{57.0} \\
      \bottomrule
      \end{tabular}%
      \vspace{-1.5em}
    \label{tab:full_sem_0}%
  \end{table*}%

  \begin{table*}[t]
    \centering
    \tiny
    \setlength{\tabcolsep}{3.5pt}
    \setlength\extrarowheight{1pt}
    \caption{Full table (2/2) of semantic segmentation mIoU results on the PartNetE dataset. This table shows the results on 28 object categories that are unique to PartNetE and are not present in PartNet.}
      \begin{tabular}{ccc|ccc|cccccccc}
      \toprule
      \multicolumn{1}{c|}{\multirow{56}[0]{*}{\begin{sideways}Non-Overlapping Categories (27)\end{sideways}}} & \multicolumn{2}{c|}{} & \multicolumn{3}{c|}{Few-shot w/ additional data (45x8+28k)} & \multicolumn{6}{c}{Few-shot (45x8)} \\
  \cmidrule{2-14}    \multicolumn{1}{c|}{} & Category & Part  & \multicolumn{1}{c}{PointNet++~\cite{qi2017pointnet++}} & \multicolumn{1}{c}{PointNext~\cite{qian2022pointnext}} & \multicolumn{1}{c|}{SoftGroup~\cite{vu2022softgroup}} & \multicolumn{1}{c}{PointNet++~\cite{qi2017pointnet++}} & \multicolumn{1}{c}{PointNext~\cite{qian2022pointnext}} & \multicolumn{1}{c}{SoftGroup~\cite{vu2022softgroup}} & \multicolumn{1}{c}{ACD~\cite{gadelha2020label}} & \multicolumn{1}{c}{Prototype~\cite{zhao2021few}} & \multicolumn{1}{c}{PartSLIP~\cite{liu2023partslip}} & \multicolumn{1}{c}{PartSLIP*~\cite{liu2023partslip}} & \multicolumn{1}{c}{Ours} \\
  \cmidrule{2-14}    \multicolumn{1}{c|}{} & Box   & lid   & 18.6  & 84.2  & 8.8   & 24.5  & 69.4  & 24.1  & 21.1  & 68.8  & 84.5 & 77.9 & \textbf{85.5} \\
  \cline{2-14}    \multicolumn{1}{c|}{} & Bucket & handle & 0.0   & 4.1   & 25.0  & 0.0   & 0.0   & 18.9  & 0.0   & 31.3  & 36.5 & 21.0 & \textbf{85.5} \\
  \cline{2-14}    \multicolumn{1}{c|}{} & \multirow{2}[0]{*}{Camera} & button & 0.0   & 0.0   & 12.6  & 0.0   & 0.0   & 13.9  & 0.0   & 6.0   & 43.2 & 45.6 & \textbf{47.6} \\
      \multicolumn{1}{c|}{} &       & lens  & 13.0  & 66.4  & 34.6  & 19.4  & 51.9  & 43.3  & 20.2  & 58.0  & 73.4 & \textbf{78.9} & \textbf{78.9} \\
  \cline{2-14}    \multicolumn{1}{c|}{} & Cart  & wheel & 6.4   & 36.3  & 23.9  & 11.6  & 47.7  & 40.8  & 31.5  & 36.8  & \textbf{88.1} & 78.7 & 84.9 \\
  \cline{2-14}    \multicolumn{1}{c|}{} & \multirow{4}[0]{*}{CoffeeMachine} & button & \textbf{32.6} & 0.0   & 2.4   & 0.0   & 0.0   & 4.3   & 0.0   & 0.7   & 6.4 & 5.7 & 5.7 \\
      \multicolumn{1}{c|}{} &       & container & 29.0  & 25.8  & 4.6   & 7.6   & 23.0  & 25.5  & 2.8   & 25.9  & 51.1 & \textbf{55.0} & 52.8 \\
      \multicolumn{1}{c|}{} &       & knob  & \textbf{32.6} & 3.6   & 8.2   & 0.0   & 0.0   & 1.3   & 0.0   & 7.8   & 32.6 & 29.6 & 31.1 \\
      \multicolumn{1}{c|}{} &       & lid   & 44.0  & 42.3  & 17.8  & 11.2  & 45.0  & 27.6  & 0.0   & 45.7  & 61.2 & 60.0 & \textbf{65.6} \\
  \cline{2-14}    \multicolumn{1}{c|}{} & \multirow{2}[0]{*}{Dispenser} & head  & 18.0  & 20.7  & 18.3  & 6.9   & 34.1  & 42.8  & 22.0  & 45.2  & \textbf{60.4} & 55.5 & 58.0 \\
      \multicolumn{1}{c|}{} &       & lid   & 6.1   & 31.2  & 19.5  & 7.0   & 11.0  & 43.0  & 16.7  & 61.6  & \textbf{87.1} & 86.4 & 86.0 \\
  \cline{2-14}    \multicolumn{1}{c|}{} & \multirow{2}[0]{*}{Eyeglasses} & body  & 77.2  & 93.0  & 77.8  & 85.8  & \textbf{94.1} & 74.5  & 82.6  & 81.7  & 84.8 & 89.0 & 86.5 \\
      \multicolumn{1}{c|}{} &       & leg   & 75.1  & 83.2  & 67.0  & 71.8  & 84.6  & 70.9  & 73.7  & 74.0  & \textbf{91.7} & 90.2 & 90.0 \\
  \cline{2-14}    \multicolumn{1}{c|}{} & FoldingChair & seat  & 10.9  & \textbf{96.4} & 14.7  & 63.4  & 94.9  & 89.0  & 74.2  & 91.2  & 86.3 & 83.6 & 89.9 \\
  \cline{2-14}    \multicolumn{1}{c|}{} & Globe & sphere & 46.5  & 92.3  & 59.0  & 51.4  & 88.8  & 85.1  & 69.8  & 88.3  & 95.7 & 92.8 & \textbf{96.5} \\
  \cline{2-14}    \multicolumn{1}{c|}{} & \multirow{3}[0]{*}{Kettle} & lid   & 16.2  & 24.5  & 46.9  & 21.4  & 54.7  & 60.2  & 22.9  & 58.9  & 78.8 & 72.1 & \textbf{85.1} \\
      \multicolumn{1}{c|}{} &       & handle & 16.2  & 71.3  & 56.8  & 33.8  & 73.1  & 60.1  & 43.7  & 73.6 & 73.5 & 70.2 & \textbf{89.4} \\
      \multicolumn{1}{c|}{} &       & spout & 30.2  & 39.6  & 68.5  & 30.5  & 53.7  & 61.8  & 54.0  & 55.5  & 78.6 & 78.0 & \textbf{82.5} \\
  \cline{2-14}    \multicolumn{1}{c|}{} & \multirow{2}[0]{*}{KitchenPot} & lid   & 25.9  & 79.6  & 49.1  & 44.1  & 80.1 & 66.8  & 69.9  & 76.1 & 77.7 & 77.6 & \textbf{82.4} \\
      \multicolumn{1}{c|}{} &       & handle & 5.7   & 34.3  & 41.9  & 19.3  & 51.8  & 42.7  & 33.8  & 50.5  & 61.5 & 56.2 & \textbf{63.4} \\
  \cline{2-14}    \multicolumn{1}{c|}{} & \multirow{3}[0]{*}{Lighter} & lid   & 52.4  & 38.4  & 32.0  & 33.6  & 39.9  & 40.5  & 32.3  & 42.8 & 69.8 & 69.9 & \textbf{73.1} \\
      \multicolumn{1}{c|}{} &       & wheel & 15.0  & 10.5  & 24.3  & 0.8   & 0.0   & 35.3  & 0.0   & 15.4  & 57.9 & 51.3 & \textbf{60.5} \\
      \multicolumn{1}{c|}{} &       & button & 37.6  & 0.0   & 34.2  & 0.0   & 0.0   & 43.7  & 0.0   & 34.0  & \textbf{66.3} & 57.4 & 65.0 \\
  \cline{2-14}    \multicolumn{1}{c|}{} & \multirow{3}[0]{*}{Mouse} & button & 3.0   & 0.8   & 20.2 & 0.0   & 2.7   & 4.8   & 0.0   & 0.1   & 16.2 & 16.0 & \textbf{21.5} \\
      \multicolumn{1}{c|}{} &       & cord  & 33.3  & 65.0  & 41.0  & 0.0   & 0.0   & 53.2  & 0.0   & 40.7  & \textbf{66.5} & 65.8 & 66.2 \\
      \multicolumn{1}{c|}{} &       & wheel & 0.0   & 0.0   & \textbf{70.8} & 0.0   & 0.0   & 31.9  & 0.0   & 19.4  & 49.4 & 47.1 & 52.1 \\
  \cline{2-14}    \multicolumn{1}{c|}{} & \multirow{2}[0]{*}{Oven} & door  & 32.3  & \textbf{75.6} & 17.2  & 38.9  & 73.5  & 49.7  & 17.8  & 68.3  & 73.1 & 73.0 & 73.2 \\
      \multicolumn{1}{c|}{} &       & knob  & 36.4  & 0.0   & 10.1  & 0.0   & 0.0   & 21.5  & 0.0   & 4.7   & \textbf{73.9} & 66.3 & 67.3 \\
  \cline{2-14}    \multicolumn{1}{c|}{} & \multirow{2}[0]{*}{Pen} & cap   & 42.7  & 53.3  & 26.3  & 8.8   & 45.4  & 40.5  & 10.8  & 34.0  & \textbf{68.4} & 68.0 & 64.4 \\
      \multicolumn{1}{c|}{} &       & button & 50.3  & 25.6  & 31.4  & 0.0   & 21.0  & 52.1  & 0.0   & 61.0  & \textbf{74.6} & 70.1 & 68.1 \\
  \cline{2-14}    \multicolumn{1}{c|}{} & \multirow{2}[0]{*}{Phone} & lid   & 40.0  & 78.7 & 0.3   & 10.3  & 66.7  & 2.0   & 19.7  & 68.3  & 74.0 & 72.3 & \textbf{86.4} \\
      \multicolumn{1}{c|}{} &       & button & 0.0   & 0.2   & 4.4   & 0.0   & 0.0   & 8.2   & 0.0   & 2.6   & 22.8 & 30.9 & \textbf{31.5} \\
  \cline{2-14}    \multicolumn{1}{c|}{} & Pliers & leg   & 57.7  & \textbf{99.6} & 74.2  & 99.3  & \textbf{99.6} & 91.2  & 83.5  & 91.0  & 33.2 & 48.1 & 29.7 \\
  \cline{2-14}    \multicolumn{1}{c|}{} & Printer & button & 0.0   & 0.0   & 1.2   & 0.0   & 0.0   & 1.6   & 0.0   & 0.2   & 4.3 & 3.2 & \textbf{6.2} \\
  \cline{2-14}    \multicolumn{1}{c|}{} & Remote & button & 3.6   & \textbf{57.8} & 37.1  & 0.0   & 0.5   & 37.5  & 0.0   & 29.6  & 38.3 & 36.0 & 36.4 \\
  \cline{2-14}    \multicolumn{1}{c|}{} & \multirow{3}[0]{*}{Safe} & door  & 14.0  & \textbf{76.7} & 9.8   & 32.7  & 67.0  & 24.8  & 28.0  & 51.9  & 64.5 & 66.3 & 71.6 \\
      \multicolumn{1}{c|}{} &       & switch & 13.6  & 0.0   & 5.8   & 0.0   & 0.0   & 21.7  & 0.0   & 5.8   & 27.9 & 34.0 & \textbf{35.3} \\
      \multicolumn{1}{c|}{} &       & button & \textbf{68.2} & 0.0   & 0.4   & 0.0   & 0.0   & 0.0   & 0.0   & 2.7   & 4.1  & 3.2 & 4.8 \\
  \cline{2-14}    \multicolumn{1}{c|}{} & \multirow{2}[0]{*}{Stapler} & body  & 58.3  & 91.4  & 83.4  & 30.4  & 91.1  & 83.9  & 49.8  & 83.0  & \textbf{93.6} & 86.4 & 86.3 \\
      \multicolumn{1}{c|}{} &       & lid   & 44.9  & \textbf{85.7} & 76.8  & 45.7  & 83.3  & 80.5  & 50.2  & 78.4  & 76.0 & 69.2 & 39.6 \\
  \cline{2-14}    \multicolumn{1}{c|}{} & \multirow{2}[0]{*}{Suitcase} & handle & 6.3   & 9.3   & 30.0  & 6.7   & 28.9  & 30.7  & 26.4  & 38.9  & 84.1 & 74.7 & \textbf{87.3} \\
      \multicolumn{1}{c|}{} &       & wheel & \textbf{75.0} & 17.8  & 6.6   & 0.0   & 0.0   & 28.9  & 0.0   & 32.1  & 56.7 & 50.7 & 52.7 \\
  \cline{2-14}    \multicolumn{1}{c|}{} & Switch & switch & 1.8   & 39.7  & 21.0  & 9.3   & 42.9  & 31.8  & 10.3  & 40.9  & \textbf{59.4} & 50.7 & 56.1 \\
  \cline{2-14}    \multicolumn{1}{c|}{} & \multirow{2}[0]{*}{Toaster} & button & 23.5  & 2.7   & 36.6  & 0.0   & 0.0   & 17.7  & 0.0   & 9.0   & \textbf{58.7} & 53.4 & 50.5 \\
      \multicolumn{1}{c|}{} &       & slider & 5.9   & 14.0  & 16.2  & 0.0   & 0.0   & 11.8  & 0.0   & 11.2  & \textbf{61.3} & 50.1 & 51.0 \\
  \cline{2-14}    \multicolumn{1}{c|}{} & \multirow{3}[0]{*}{Toilet} & lid   & 19.5  & 49.4  & 12.7  & 9.4   & 68.5  & 27.9  & 53.4  & 56.8  & 72.6 & 68.7 & \textbf{75.9} \\
      \multicolumn{1}{c|}{} &       & seat  & \textbf{62.3} & 0.0   & 2.9   & 0.0   & 0.0   & 6.2   & 0.0   & 0.1   & 21.3 & 23.5 & 29.7 \\
      \multicolumn{1}{c|}{} &       & button & 16.4  & 0.0   & 23.2  & 0.0   & 0.0   & 7.6   & 0.0   & 1.6   & \textbf{67.6} & 53.3 & 65.1 \\
  \cline{2-14}    \multicolumn{1}{c|}{} & \multirow{2}[0]{*}{USB} & cap   & 54.9  & 67.2  & 61.6  & 21.1  & \textbf{79.7} & 73.9  & 11.4  & 72.6  & 58.1 & 55.1 & 55.1 \\
      \multicolumn{1}{c|}{} &       & rotation & 49.8  & \textbf{68.6} & 26.6  & 35.7  & 61.7  & 38.1  & 38.9  & 58.1  & 50.7 & 57.8 & 59.8 \\
  \cline{2-14}    \multicolumn{1}{c|}{} & \multirow{2}[0]{*}{WashingMachine} & door  & 1.1   & 54.5  & 25.8  & 8.9   & 37.9  & 40.0  & 20.2  & 55.4  & 63.3 & 59.0 & \textbf{64.9} \\
      \multicolumn{1}{c|}{} &       & button & 0.0   & 0.0   & 22.4  & 0.0   & 0.0   & 5.0   & 0.0   & 6.7   & \textbf{43.6} & 30.8 & 32.5 \\
  \cline{2-14}    \multicolumn{1}{c|}{} & Window & window & 26.3  & \textbf{83.3} & 39.2  & 62.6  & 83.2  & 66.4  & 66.8  & 76.6  & 75.4 & 78.7 & 72.8 \\
  \cmidrule{2-14}    \multicolumn{1}{c|}{} & \multicolumn{2}{c|}{Overall (28)} & 25.4  & \textbf{45.1}  & 30.7  & 21.8  & 41.5  & 41.1  & 25.6  & 46.3  & 61.3 & 58.7 & \textbf{63.3} \\
      \midrule
      \multicolumn{3}{c|}{Overall (45)} & 36.8  & \textbf{50.2}  & 38.1  & 20.4  & 40.6  & 38.0  & 23.2  & 44.3  & 59.4 & 57.9 & \textbf{60.8} \\
      \bottomrule
      \end{tabular}%
      \vspace{-1.5em}
    \label{tab:full_sem_1}%
  \end{table*}%
  
\begin{table*}[t]
    \centering
    \tiny
    \setlength{\tabcolsep}{1.5pt}
    \setlength\extrarowheight{1pt}
    \caption{Full table of instance segmentation mAP@50 results on the PartNetE dataset.}
      \begin{tabular}{cccccccccc|ccc|cc|ccccc}
      \toprule
      \multicolumn{1}{c|}{\multirow{54}[0]{*}{\begin{sideways}Overlapping Categories\end{sideways}}} & \multicolumn{1}{c}{\multirow{3}[0]{*}{Category}} & \multicolumn{1}{c|}{\multirow{3}[0]{*}{Part}} & \multicolumn{2}{c|}{45x8+28k} & \multicolumn{5}{c|}{Few-shot (45x8)} & \multicolumn{1}{c|}{\multirow{57}[0]{*}{\begin{sideways}Non-Overlapping Categories\end{sideways}}} & \multirow{3}[0]{*}{Category} & \multirow{3}[0]{*}{Part} & \multicolumn{2}{c|}{45x8+28k} & \multicolumn{5}{c}{Few-shot (45x8)} \\
  \cline{4-10}\cline{14-20}    \multicolumn{1}{c|}{} &       & \multicolumn{1}{c|}{} & \multicolumn{1}{c}{Point} & \multicolumn{1}{c|}{Soft} & \multicolumn{1}{c}{Point} & \multicolumn{1}{c}{Soft} & \multicolumn{1}{c}{Part} & \multicolumn{1}{c}{Part} & \multicolumn{1}{c|}{\multirow{2}[0]{*}{Ours}} & \multicolumn{1}{c|}{} &       &       & \multicolumn{1}{c}{Point} & \multicolumn{1}{c|}{Soft} & \multicolumn{1}{c}{Point} & \multicolumn{1}{c}{Soft} & \multicolumn{1}{c}{Part} & \multicolumn{1}{c}{Part} & \multicolumn{1}{c}{\multirow{2}[0]{*}{Ours}} \\
      \multicolumn{1}{c|}{} &       & \multicolumn{1}{c|}{} & \multicolumn{1}{c}{Group~\cite{jiang2020pointgroup}} & \multicolumn{1}{c|}{Group~\cite{vu2022softgroup}} & \multicolumn{1}{c}{Group~\cite{jiang2020pointgroup}} & \multicolumn{1}{c}{Group~\cite{vu2022softgroup}} & \multicolumn{1}{c}{SLIP~\cite{liu2023partslip}} & \multicolumn{1}{c}{SLIP*~\cite{liu2023partslip}}  &    & \multicolumn{1}{c|}{} &       &       & \multicolumn{1}{c}{Group~\cite{jiang2020pointgroup}} & \multicolumn{1}{c|}{Group~\cite{vu2022softgroup}} & \multicolumn{1}{c}{Group~\cite{jiang2020pointgroup}} & \multicolumn{1}{c}{Group~\cite{vu2022softgroup}} & \multicolumn{1}{c}{SLIP~\cite{liu2023partslip}} & \multicolumn{1}{c}{SLIP*~\cite{liu2023partslip}} &  \\
  \cline{2-10}\cline{12-20}    \multicolumn{1}{c|}{} & \multicolumn{1}{c}{Bottle} & \multicolumn{1}{c|}{lid} & 38.2  & \multicolumn{1}{c|}{43.9} & 8.0   & 22.4  & \textbf{79.4} & 74.4 & 78.5 & \multicolumn{1}{c|}{} & Box   & lid   & 7.2   & 8.6   & 15.8  & 19.7  & \textbf{77.2} & 55.3 & 65.4 \\
  \cline{2-10}\cline{12-20}    \multicolumn{1}{c|}{} & \multicolumn{1}{c}{\multirow{5}[0]{*}{Chair}} & \multicolumn{1}{c|}{arm} & 94.6  & \multicolumn{1}{c|}{\textbf{95.1}} & 35.9  & 71.0  & \multicolumn{1}{c}{67.7} & 51.7 & 64.9 & \multicolumn{1}{c|}{} & Bucket & handle & 1.5   & 1.6   & 1.0   & 1.1   & 18.2 & 16.7 & \textbf{87.5} \\
  \cline{12-20}    \multicolumn{1}{c|}{} &       & \multicolumn{1}{c|}{back} & 82.0  & \multicolumn{1}{c|}{73.2} & 83.8  & 93.7  & \multicolumn{1}{c}{\textbf{95.4}} & 88.4 & 88.9 & \multicolumn{1}{c|}{} & \multirow{2}[0]{*}{Camera} & button & 1.0   & 1.5   & 4.5   & 6.1   & 33.8 & 25.2 & \textbf{35.2} \\
      \multicolumn{1}{c|}{} &       & \multicolumn{1}{c|}{leg} & 88.6  & \multicolumn{1}{c|}{\textbf{93.6}} & 92.2  & 89.9  & \multicolumn{1}{c}{78.1} & 74.1 & 86.5 & \multicolumn{1}{c|}{} &       & lens  & 16.1  & 0.0   & 5.0   & 16.4  & \textbf{39.9} & 34.0 & 34.1 \\
  \cline{12-20}    \multicolumn{1}{c|}{} &       & \multicolumn{1}{c|}{seat} & 75.0  & \multicolumn{1}{c|}{85.9} & 81.4  & 88.1  & \multicolumn{1}{c}{85.5} & 89.9 & \textbf{92.9} & \multicolumn{1}{c|}{} & Cart  & wheel & 29.2  & 28.4  & 28.5  & 29.8  & \textbf{83.3} & 71.0 & 80.8 \\
  \cline{12-20}    \multicolumn{1}{c|}{} &       & \multicolumn{1}{c|}{wheel} & 98.0  & \multicolumn{1}{c|}{\textbf{97.7}} & 92.8  & 95.9  & \multicolumn{1}{c}{95.5} & 92.4 & 97.0 & \multicolumn{1}{c|}{} & \multirow{4}[0]{*}{CoffeeMachine} & button & 1.0   & 1.0   & 1.1   & 0.0   & \textbf{2.2} & 1.4 & 1.5 \\
  \cline{2-10}    \multicolumn{1}{c|}{} & \multicolumn{1}{c}{Clock} & \multicolumn{1}{c|}{hand} & 1.0   & \multicolumn{1}{c|}{1.0} & 1.0   & 1.0   & \multicolumn{1}{c}{14.9} & 25.9 & \textbf{39.0} & \multicolumn{1}{c|}{} &       & container & 2.5   & 4.0   & 13.6  & 19.7  & \textbf{32.8} & 21.5 & 20.5 \\
  \cline{2-10}    \multicolumn{1}{c|}{} & \multicolumn{1}{c}{\multirow{2}[0]{*}{Dishwasher}} & \multicolumn{1}{c|}{door} & \textbf{76.7} & \multicolumn{1}{c|}{75.0} & 50.6  & 55.6  & \multicolumn{1}{c}{57.4} & 49.1 & 57.2 & \multicolumn{1}{c|}{} &       & knob  & 5.6   & 5.0   & 3.3   & 1.5   & 13.5 & 16.0 & \textbf{14.7} \\
      \multicolumn{1}{c|}{} &       & \multicolumn{1}{c|}{handle} & 55.6  & \multicolumn{1}{c|}{\textbf{56.4}} & 1.0   & 26.4  & \multicolumn{1}{c}{32.9} & 30.5 & 31.8 & \multicolumn{1}{c|}{} &       & lid   & 3.3   & 1.4   & 8.9   & 22.6  & \textbf{27.6} & 23.9 & 18.4 \\
  \cline{2-10}\cline{12-20}    \multicolumn{1}{c|}{} & \multicolumn{1}{c}{\multirow{3}[0]{*}{Display}} & \multicolumn{1}{c|}{base} & 95.2  & \multicolumn{1}{c|}{\textbf{97.4}} & 13.2  & 22.1  & \multicolumn{1}{c}{94.2} & 94.1 & 95.6 & \multicolumn{1}{c|}{} & \multirow{2}[0]{*}{Dispenser} & head  & 27.5  & 29.2  & 39.1  & 45.4  & \textbf{46.4} & 40.1 & 41.4 \\
      \multicolumn{1}{c|}{} &       & \multicolumn{1}{c|}{screen} & 46.0  & \multicolumn{1}{c|}{55.4} & 32.9  & 49.2  & \multicolumn{1}{c}{\textbf{70.7}} & 52.5 & 70.6 & \multicolumn{1}{c|}{} &       & lid   & 20.5  & 23.6  & 22.4  & 30.2  & 80.6 & 79.3 & \textbf{85.1} \\
  \cline{12-20}    \multicolumn{1}{c|}{} &       & \multicolumn{1}{c|}{support} & 54.0  & \multicolumn{1}{c|}{53.2} & 4.1   & 11.1  & \multicolumn{1}{c}{\textbf{84.0}} & 68.0 & 56.0 & \multicolumn{1}{c|}{} & \multirow{2}[0]{*}{Eyeglasses} & body  & 31.7  & 39.5  & 28.1  & 34.7  & \textbf{79.5} & 54.1 & 57.8 \\
  \cline{2-10}    \multicolumn{1}{c|}{} & \multicolumn{1}{c}{\multirow{3}[0]{*}{Door}} & \multicolumn{1}{c|}{frame} & \textbf{36.8} & \multicolumn{1}{c|}{28.3} & 2.7   & 9.8   & \multicolumn{1}{c}{2.8} & 3.1 & 3.0 & \multicolumn{1}{c|}{} &       & leg   & 68.0  & 62.7  & 50.3  & 56.3  & \textbf{84.9} & 79.9 & 83.1 \\
  \cline{12-20}    \multicolumn{1}{c|}{} &       & \multicolumn{1}{c|}{door} & 32.4  & \multicolumn{1}{c|}{\textbf{34.3}} & 7.5   & 5.9   & \multicolumn{1}{c}{30.7} & 20.5 & 26.3  & \multicolumn{1}{c|}{} & FoldingChair & seat  & 16.8  & 16.8  & 86.4  & 79.0  & 76.7 & 75.6 & \textbf{81.9} \\
  \cline{12-20}    \multicolumn{1}{c|}{} &       & \multicolumn{1}{c|}{handle} & 1.0   & \multicolumn{1}{c|}{1.0} & 1.0   & 1.0   & \multicolumn{1}{c}{20.3} & 18.4 & \textbf{23.8}  & \multicolumn{1}{c|}{} & Globe & sphere & 63.1  & 63.1  & 80.2  & 75.7  & 81.0 & 80.8 & \textbf{85.4} \\
  \cline{2-10}\cline{12-20}    \multicolumn{1}{c|}{} & \multicolumn{1}{c}{\multirow{2}[0]{*}{Faucet}} & \multicolumn{1}{c|}{spout} & 85.4  & \multicolumn{1}{c|}{\textbf{86.3}} & 50.7  & 52.4  & \multicolumn{1}{c}{61.7} & 60.8 & 61.8 & \multicolumn{1}{c|}{} & \multirow{3}[0]{*}{Kettle} & lid   & 64.0  & 64.4  & 65.8  & 70.0  & 76.1 & 73.2 & \textbf{91.7} \\
      \multicolumn{1}{c|}{} &       & \multicolumn{1}{c|}{switch} & \textbf{74.5} & \multicolumn{1}{c|}{72.5} & 11.2  & 22.2  & \multicolumn{1}{c}{47.6} & 36.4 & 31.0  & \multicolumn{1}{c|}{} &       & handle & 51.4  & 54.3  & 45.0  & 59.0  & \textbf{78.1} & 74.2 & 74.5 \\
  \cline{2-10}    \multicolumn{1}{c|}{} & \multicolumn{1}{c}{\multirow{2}[0]{*}{Keyboard}} & \multicolumn{1}{c|}{cord} & 42.6  & \multicolumn{1}{c|}{39.7} & 34.3  & 21.3  & \multicolumn{1}{c}{68.6} & 86.0 & \textbf{86.1} & \multicolumn{1}{c|}{} &       & spout & 68.5  & 72.6 & 45.4  & 61.3  & 71.9 & 70.0  & \textbf{78.1} \\
  \cline{12-20}    \multicolumn{1}{c|}{} &       & \multicolumn{1}{c|}{key} & 37.2  & \multicolumn{1}{c|}{37.7} & 16.1  & 1.0   & \multicolumn{1}{c}{12.3} &  34.1 & \textbf{40.2} & \multicolumn{1}{c|}{} & \multirow{2}[0]{*}{KitchenPot} & lid   & 68.3  & 68.5  & 81.4  & 87.1  & 91.5 & 91.1 & \textbf{91.9} \\
  \cline{2-10}    \multicolumn{1}{c|}{} & \multicolumn{1}{c}{Knife} & \multicolumn{1}{c|}{blade} & 19.3  & \multicolumn{1}{c|}{27.2} & 15.6  & 10.3  & \multicolumn{1}{c}{43.9} & 43.3 & \textbf{46.0} & \multicolumn{1}{c|}{} &       & handle & 50.6 & 50.1  & 32.5  & 44.3  & 49.5 & 49.5  & \textbf{69.6} \\
  \cline{2-10}\cline{12-20}    \multicolumn{1}{c|}{} & \multicolumn{1}{c}{\multirow{4}[0]{*}{Lamp}} & \multicolumn{1}{c|}{base} & 64.3  & \multicolumn{1}{c|}{71.1} & 8.5   & 17.9  & \multicolumn{1}{c}{\textbf{89.9}} & 87.5 & 88.6 & \multicolumn{1}{c|}{} & \multirow{3}[0]{*}{Lighter} & lid   & 30.7  & 30.7  & 0.0   & 40.6  & 45.8 & 50.8 & \textbf{51.6} \\
      \multicolumn{1}{c|}{} &       & \multicolumn{1}{c|}{body} & 48.6  & \multicolumn{1}{c|}{36.5} & 4.3   & 11.0  & \multicolumn{1}{c}{\textbf{87.4}} & 86.8 & 84.1  & \multicolumn{1}{c|}{} &       & wheel & 6.0   & 5.3   & 0.0   & 47.9 & 34.3  & 32.3 & \textbf{48.4} \\
      \multicolumn{1}{c|}{} &       & \multicolumn{1}{c|}{bulb} & 54.5  & \multicolumn{1}{c|}{\textbf{59.2}} & 7.1   & 1.9   & \multicolumn{1}{c}{5.9} & 14.9 & 9.3  & \multicolumn{1}{c|}{} &       & button & 64.1  & \textbf{67.8} & 0.0   & 63.2  & 23.6 & 28.1 & 27.1 \\
  \cline{12-20}    \multicolumn{1}{c|}{} &       & \multicolumn{1}{c|}{shade} & 83.5  & \multicolumn{1}{c|}{86.4} & 19.4  & 47.0  & \multicolumn{1}{c}{\textbf{90.1}} & 88.9 & 85.5 & \multicolumn{1}{c|}{} & \multirow{3}[0]{*}{Mouse} & button & 1.0   & 1.0   & 0.0   & 0.0   & 1.7 & \textbf{2.3} & 2.0 \\
  \cline{2-10}    \multicolumn{1}{c|}{} & \multicolumn{1}{c}{\multirow{5}[0]{*}{Laptop}} & \multicolumn{1}{c|}{keyboard} & 0.0   & \multicolumn{1}{c|}{0.0} & 40.1  & 53.8 & \multicolumn{1}{c}{53.4} & 51.5 & \textbf{75.5} & \multicolumn{1}{c|}{} &       & cord  & 1.0   & 1.0   & 0.0   & 1.0   & \textbf{66.3} & \textbf{66.3} & \textbf{66.3} \\
      \multicolumn{1}{c|}{} &       & \multicolumn{1}{c|}{screen} & 1.0   & \multicolumn{1}{c|}{1.0} & 36.3  & \textbf{61.5} & \multicolumn{1}{c}{48.5} & 32.0 & 55.7 & \multicolumn{1}{c|}{} &       & wheel & \textbf{83.2} & \textbf{83.2} & 0.0   & 53.7  & 50.5 & 42.0 & 49.3 \\
  \cline{12-20}    \multicolumn{1}{c|}{} &       & \multicolumn{1}{c|}{shaft} & 1.2   & \multicolumn{1}{c|}{3.5} & 1.0   & 0.0   & \multicolumn{1}{c}{2.0} & 1.4 & \textbf{4.0} & \multicolumn{1}{c|}{} & \multirow{2}[0]{*}{Oven} & door  & 26.5  & 31.9  & 0.0   & 19.1  & \textbf{54.9} & 47.4 & 44.6 \\
      \multicolumn{1}{c|}{} &       & \multicolumn{1}{c|}{touchpad} & 0.0   & \multicolumn{1}{c|}{0.0} & 0.0   & 0.0   & \multicolumn{1}{c}{\textbf{19.7}} & 12.9 & 11.1 & \multicolumn{1}{c|}{} &       & knob  & 1.0   & 1.0   & 0.0   & 1.6   & \textbf{74.1} & 45.2 & 68.0 \\
  \cline{12-20}    \multicolumn{1}{c|}{} &       & \multicolumn{1}{c|}{camera} & 0.0   & \multicolumn{1}{c|}{0.0} & 0.0   & 0.0   & \multicolumn{1}{c}{\textbf{1.0}} & \textbf{1.0} & \textbf{1.0} & \multicolumn{1}{c|}{} & \multirow{2}[0]{*}{Pen} & cap   & 48.2  & 44.4  & 0.0   & 44.3  & \textbf{51.6} & 41.2 & 34.2 \\
  \cline{2-10}    \multicolumn{1}{c|}{} & \multicolumn{1}{c}{\multirow{4}[0]{*}{Microwave}} & \multicolumn{1}{c|}{display} & 4.2   & \multicolumn{1}{c|}{1.0} & 0.0   & 1.0   & \multicolumn{1}{c}{6.3} & \textbf{25.2} & 20.6 & \multicolumn{1}{c|}{} &       & button & 16.9  & 16.9  & 0.0   & 10.9  & 37.9 & 44.6 & \textbf{46.2} \\
  \cline{12-20}    \multicolumn{1}{c|}{} &       & \multicolumn{1}{c|}{door} & 62.6 & \multicolumn{1}{c|}{57.1} & 0.0   & 31.0  & \multicolumn{1}{c}{34.4} & 40.9 & \textbf{63.9} & \multicolumn{1}{c|}{} & \multirow{2}[0]{*}{Phone} & lid   & 1.0   & 1.1   & 0.0   & 1.2   & 37.8 & \textbf{50.8} & 40.7 \\
      \multicolumn{1}{c|}{} &       & \multicolumn{1}{c|}{handle} & 1.0   & \multicolumn{1}{c|}{1.0} & 0.0   & 0.0   & \multicolumn{1}{c}{60.4} & 50.5 & \textbf{90.2} & \multicolumn{1}{c|}{} &       & button & 1.0   & 1.0   & 0.0   & 1.0   & 26.6 & 32.7 & \textbf{33.8} \\
  \cline{12-20}    \multicolumn{1}{c|}{} &       & \multicolumn{1}{c|}{button} & \textbf{100.0} & \multicolumn{1}{c|}{\textbf{100.0}} & 0.0   & 22.8  & \multicolumn{1}{c}{3.2} & 12.1 & 5.2 & \multicolumn{1}{c|}{} & Pliers & leg   & 28.2  & \textbf{40.4} & 6.8   & 14.5  & 4.7  & 3.2 & 7.9 \\
  \cline{2-10}\cline{12-20}    \multicolumn{1}{c|}{} & \multicolumn{1}{c}{\multirow{2}[0]{*}{Refrigerator}} & \multicolumn{1}{c|}{door} & \textbf{57.1} & \multicolumn{1}{c|}{54.2} & 0.0   & 23.2  & \multicolumn{1}{c}{31.3} & 36.3 & 44.2 & \multicolumn{1}{c|}{} & Printer & button & 1.0   & 1.0   & 0.0   & 0.0   & 1.3 & 1.3 & \textbf{1.5} \\
  \cline{12-20}    \multicolumn{1}{c|}{} &       & \multicolumn{1}{c|}{handle} & 19.3  & \multicolumn{1}{c|}{17.2} & 0.0   & 9.7   & \multicolumn{1}{c}{\textbf{39.7}} & 23.3 & 36.8 & \multicolumn{1}{c|}{} & Remote & button & \textbf{23.4} & 22.5  & 0.0   & 6.2   & 23.1 & 21.1 & 21.7 \\
  \cline{2-10}\cline{12-20}    \multicolumn{1}{c|}{} & \multicolumn{1}{c}{\multirow{3}[0]{*}{Scissors}} & \multicolumn{1}{c|}{blade} & 6.2   & \multicolumn{1}{c|}{6.5} & 4.5   & 3.0   & \multicolumn{1}{c}{14.1} & 7.4 & 28.2 & \multicolumn{1}{c|}{} & \multirow{3}[0]{*}{Safe} & door  & 11.0  & 12.3  & 0.0   & 19.4  & 68.4 & 60.0 & \textbf{69.3} \\
      \multicolumn{1}{c|}{} &       & \multicolumn{1}{c|}{handle} & 82.0  & \multicolumn{1}{c|}{\textbf{82.9}} & 41.9  & 34.5  & \multicolumn{1}{c}{58.4} & 44.0 & 77.0 & \multicolumn{1}{c|}{} &       & switch & 4.8   & 5.4   & 0.0   & 23.3  & \textbf{27.4} & 15.6 & 25.2 \\
      \multicolumn{1}{c|}{} &       & \multicolumn{1}{c|}{screw} & 27.2  & \multicolumn{1}{c|}{\textbf{28.4}} & 8.9   & 4.6   & \multicolumn{1}{c}{4.3} & 3.0 & 7.4 & \multicolumn{1}{c|}{} &       & button & \textbf{1.0} & \textbf{1.0} & 0.0   & \textbf{1.0} & \textbf{1.0} & \textbf{1.0} & \textbf{1.0} \\
  \cline{2-10}\cline{12-20}    \multicolumn{1}{c|}{} & \multicolumn{1}{c}{\multirow{3}[0]{*}{StorageFurniture}} & \multicolumn{1}{c|}{door} & \textbf{86.9} & \multicolumn{1}{c|}{85.6} & 0.0   & 28.8  & \multicolumn{1}{c}{24.9} & 20.2 & 29.1 & \multicolumn{1}{c|}{} & \multirow{2}[0]{*}{Stapler} & body  & 86.6  & 96.7  & 52.4  & 88.0  & \textbf{100.0} & 89.2 & 91.9 \\
      \multicolumn{1}{c|}{} &       & \multicolumn{1}{c|}{drawer} & 3.9   & \multicolumn{1}{c|}{4.2} & 0.0   & 1.5   & \multicolumn{1}{c}{6.1} & 4.4 & \textbf{10.6} & \multicolumn{1}{c|}{} &       & lid   & 90.0  & \textbf{91.8} & 69.8  & 78.2  & 89.7  & 58 & 78.3 \\
  \cline{12-20}    \multicolumn{1}{c|}{} &       & \multicolumn{1}{c|}{handle} & 56.4  & \multicolumn{1}{c|}{57.5} & 0.0   & 4.6   & \multicolumn{1}{c}{67.5} & 63.0 & \textbf{72.8} & \multicolumn{1}{c|}{} & \multirow{2}[0]{*}{Suitcase} & handle & 25.5  & 24.2  & 0.0   & 12.9  & 64.1 & 63.9 & \textbf{69.3} \\
  \cline{2-10}    \multicolumn{1}{c|}{} & \multicolumn{1}{c}{\multirow{6}[0]{*}{Table}} & \multicolumn{1}{c|}{door} & 44.4  & \multicolumn{1}{c|}{\textbf{49.3}} & 0.0   & 0.0   & \multicolumn{1}{c}{0.0} & 0.0  & 0.0 & \multicolumn{1}{c|}{} &       & wheel & 5.7   & 2.9   & 0.0   & 3.1   & 25.7 & 25.3  & \textbf{29.9} \\
  \cline{12-20}    \multicolumn{1}{c|}{} &       & \multicolumn{1}{c|}{drawer} & 35.7  & \multicolumn{1}{c|}{\textbf{36.5}} & 0.0   & 0.0   & \multicolumn{1}{c}{11.3} & 10.9 & 17.2 & \multicolumn{1}{c|}{} & Switch & switch & 7.5   & 5.6   & 0.0   & 21.2  & \textbf{35.1} & 24.6 & 26.8 \\
  \cline{12-20}    \multicolumn{1}{c|}{} &       & \multicolumn{1}{c|}{leg} & 33.8  & \multicolumn{1}{c|}{27.4} & 0.0   & 7.7   & \multicolumn{1}{c}{45.9} & 45.7 & \textbf{50.0} & \multicolumn{1}{c|}{} & \multirow{2}[0]{*}{Toaster} & button & 9.0   & 10.1  & 0.0   & 4.5   & \textbf{31.4} & 26.1 & 28.6 \\
      \multicolumn{1}{c|}{} &       & \multicolumn{1}{c|}{tabletop} & 81.2  & \multicolumn{1}{c|}{\textbf{82.0}} & 0.0   & 30.0  & \multicolumn{1}{c}{64.1} & 63.8 & 64.6 & \multicolumn{1}{c|}{} &       & slider & 5.0   & 5.0   & 0.0   & 16.9  & 45.4 & 43.7 & \textbf{54.6} \\
  \cline{12-20}    \multicolumn{1}{c|}{} &       & \multicolumn{1}{c|}{wheel} & 1.0   & \multicolumn{1}{c|}{1.3} & 0.0   & 1.1   & \multicolumn{1}{c}{\textbf{64.7}} & 64.5 & 54.0 & \multicolumn{1}{c|}{} & \multirow{3}[0]{*}{Toilet} & lid   & 5.5   & 6.1   & 0.0   & 37.5  & \textbf{62.3} & 42.5 & 50.0 \\
      \multicolumn{1}{c|}{} &       & \multicolumn{1}{c|}{handle} & \textbf{81.9} & \multicolumn{1}{c|}{80.8} & 0.0   & 46.4  & \multicolumn{1}{c}{7.6} & 7.6 & 15.3 & \multicolumn{1}{c|}{} &       & seat  & 0.0   & 0.0   & 0.0   & 1.0   & 4.2 & 5.4 & \textbf{10.9} \\
  \cline{2-10}    \multicolumn{1}{c|}{} & \multicolumn{1}{c}{\multirow{3}[0]{*}{TrashCan}} & \multicolumn{1}{c|}{footpedal} & 34.8  & \multicolumn{1}{c|}{\textbf{35.3}} & 0.0   & 15.3  & \multicolumn{1}{c}{0.0} & 0.0 & 0.0 & \multicolumn{1}{c|}{} &       & button & 1.0   & 1.0   & 0.0   & 1.5   & \textbf{70.3} & 69.7 & 63.1 \\
  \cline{12-20}    \multicolumn{1}{c|}{} &       & \multicolumn{1}{c|}{lid} & 0.0   & \multicolumn{1}{c|}{0.0} & 0.0   & 1.0   & \multicolumn{1}{c}{37.8} & 33.6 & \textbf{39.8} & \multicolumn{1}{c|}{} & \multirow{2}[0]{*}{USB} & cap   & 67.3  & \textbf{75.7} & 0.0   & 69.0  & 26.0 & 20.3 & 32.1 \\
      \multicolumn{1}{c|}{} &       & \multicolumn{1}{c|}{door} & 0.0   & \multicolumn{1}{c|}{0.0} & 0.0   & 1.0   & \multicolumn{1}{c}{1.0} & 1.6 & \textbf{1.9} & \multicolumn{1}{c|}{} &       & rotation & 16.3  & 15.0  & 0.0   & 33.3 & 29.7 & 26.2 & \textbf{34.1} \\
  \cline{2-10}\cline{12-20}    \multicolumn{1}{c|}{} & \multicolumn{2}{c|}{Overall (17)} & 41.7  & \multicolumn{1}{c|}{\textbf{42.4}} & 14.6  & 21.3  & \multicolumn{1}{c}{42.5} & 41.1 & \textbf{47.6} & \multicolumn{1}{c|}{} & \multirow{2}[0]{*}{WashingMachine} & door  & 25.0  & 34.3  & 0.0   & 41.5  & \textbf{46.4} & 41.4 & 45.1 \\
  \cline{1-10}          &       &       &       &       &       &    &   &       &       & \multicolumn{1}{c|}{} &       & button & 0.0   & 0.0   & 0.0   & 1.0   & \textbf{14.1} & 12.8 & 11.9 \\
  \cline{12-20}          &       &       &       &       &       &   &    &       &       & \multicolumn{1}{c|}{} & Window & window & 21.2  & \textbf{26.4} & 0.0   & 4.3   & 15.6  & 20.1 & 19.3 \\
  \cline{12-20}          &       &       &       &       &       &    &   &       &       & \multicolumn{1}{c|}{} & \multicolumn{2}{c|}{Overall (28)} & 24.6  & 25.6 & 16.8  & 28.4  & 46.2 & 39.8 & \textbf{48.2} \\
  \cline{11-20}          &       &       &       &       &       &     &  &       &       & \multicolumn{3}{c|}{Overall (45)} & 31.0  & \textbf{31.9} & 16.0  & 25.7  & 44.8 & 40.3 & \textbf{48.0} \\
  \bottomrule  
  \end{tabular}%
  \vspace{-1.5em}
    \label{tab:ins_full}%
  \end{table*}%


\end{document}